\documentclass[runningheads]{llncs}
\usepackage{graphicx}%
\usepackage[utf8]{inputenc} %
\usepackage[T1]{fontenc}    %
\usepackage{lmodern}
\usepackage{url}            %
\usepackage{booktabs}       %
\usepackage{amsfonts}       %
\usepackage{nicefrac}       %
\usepackage{microtype}      %

\usepackage{comment}
\usepackage{amsmath,amssymb} %
\usepackage[dvipsnames]{xcolor}
\usepackage{color}
\usepackage{caption}
\usepackage{subcaption}
\usepackage{url}
\usepackage{etoolbox}
\usepackage{booktabs}
\usepackage{xspace}
\usepackage{xcolor}
\usepackage[export]{adjustbox}
\usepackage{floatrow}
\usepackage{transparent}
\usepackage{overpic}
\usepackage{sidecap}
\usepackage{tikz}
\usetikzlibrary{arrows.meta}
\usepackage{centernot}

\DeclareFloatVCode{myrowsep}{\vskip 2ex}
\usepackage[colorlinks]{hyperref}

\usepackage[accsupp]{axessibility}  %

\usepackage{array}
\newcolumntype{C}[1]{>{\centering\let\newline\\\arraybackslash\hspace{0pt}}m{#1}}

\begin{document}
\pagestyle{headings}
\mainmatter
\def\ECCVSubNumber{7043}  %

\title{The Missing Link:\\ Finding label relations across datasets} %

\titlerunning{The Missing Link}
\author{Jasper Uijlings$^\star$ \and
Thomas Mensink\thanks{Equal contribution.}\and
Vittorio Ferrari}
\authorrunning{J. Uijlings et al.}
\institute{Google Research\\
{\tt \small \email{\{jrru, mensink, vittoferrari\}@google.com}}}
\maketitle

\providetoggle{showcomments}
\settoggle{showcomments}{true} 								%

\iftoggle{showcomments}{%
    \newcommand{\changed}[1]{\textcolor{blue}{#1}}
    \newcommand{\todo}[1]{\textcolor{blue}{\textbf{TODO:} #1}}
    \newcommand{\resolved}[3][]{\ifstrequal{#1}{resolved}{\textcolor{blue}{RESOLVED:}~\textbf{{\MakeUppercase #2:}}~{#3}}{\textbf{\MakeUppercase #2:}~#3}}
    \newcommand{\andrea}[2][]{\textcolor{ForestGreen}{\resolved[#1]{andrea}{#2}}}
    \newcommand{\vitto}[2][]{\textcolor{red}{\resolved[#1]{vitto}{#2}}}
    \newcommand{\jasper}[2][]{\textcolor{violet}{\resolved[#1]{jasper}{#2}}}
    \newcommand{\tm}[2][]{\textcolor{magenta}{\resolved[#1]{TM}{#2}}}
    \newcommand{\att}[1]{\textcolor{red}{#1}}
}{%
    \newcommand{\changed}[1]{#1}
    \newcommand{\todo}[1]{}
    \newcommand{\andrea}[2][]{}
    \newcommand{\vitto}[2][]{}
    \newcommand{\jasper}[2][]{}
    \newcommand{\tm}[2][]{}
    \newcommand{\att}[1]{#1}
}
\newcommand{\TM}[2][]{\tm[#1]{#2}}
\newcommand{\DONE}{\done}
\newcommand{\softiou}{SoftIoU-EEP}
\newcommand{\msleep}{MS-LEEP}
\newcommand{\ioueep}{IoU-EEP}
\newcommand{\eleep}{E-LEEP}
\newcommand{\para}[1]{
    \par\noindent\textbf{#1}
}
\newcommand{\class}[1]{{\tt #1\xspace}}
\newcommand{\ade}[1]{{\color{Blue} \tt #1\xspace}}
\newcommand{\coco}[1]{{\color{BrickRed} \tt #1\xspace}}
\newcommand{\mseg}[1]{{\color{ForestGreen} \tt #1\xspace}}
\newcommand{\bdd}[1]{{\color{Plum} \tt #1\xspace}}
\newcommand{\ilsvrc}[1]{{\color{Plum} \tt #1\xspace}}  %

\newcommand{\Ade}[1]{ADE20k \ade{#1}}
\newcommand{\Coco}[1]{COCO \coco{#1}}
\newcommand{\Mseg}[1]{MSEG \mseg{#1}}
\newcommand{\Bdd}[1]{BDD \bdd{#1}}

\DeclareRobustCommand\mytikzline{\tikz \draw[darkgray, very thick] (0,0) -- (0.3,0);}
\DeclareRobustCommand\mytikzarrow{\tikz \draw[darkgray, thick, -{Latex[round]}] (0,0) -- (0.3,0);}
\DeclareRobustCommand\mytikzdashedline{\tikz \draw[darkgray, very thick, dashed] (0,0) -- (0.3,0);}
\begin{abstract}
Computer vision is driven by the many datasets available for training or evaluating novel methods.
However, each dataset has a different set of class labels, visual definition of classes, images following a specific distribution, annotation protocols, etc.
In this paper we explore the automatic discovery of visual-semantic relations between labels across datasets.
We aim to understand how instances of a certain class in a dataset relate to the instances of another class in another dataset. Are they in an \emph{identity}, \emph{parent/child}, \emph{overlap} relation? Or is there no link between them at all?
To find relations between labels across datasets, we propose methods based on language, on vision, and on their combination.
We show that we can effectively discover label relations across datasets, as well as their type.
We apply our method to four applications: understand label relations, identify missing aspects, increase label specificity, and predict transfer learning gains.
We conclude that label relations cannot be established by looking at the names of classes alone, as they depend strongly on how each of the datasets was constructed.
\end{abstract}

\section{Introduction}

\begin{figure}[t]
\vspace{-.3cm}
    \centering
    \begin{subfigure}[b]{0.22\linewidth}
    \fbox{\includegraphics[height=2cm]{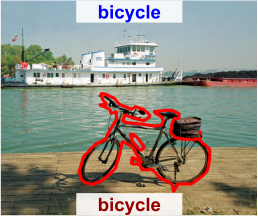}}
    \caption{identity}
    \end{subfigure}
    \begin{subfigure}[b]{0.285\linewidth}
    \fbox{\includegraphics[height=2cm]{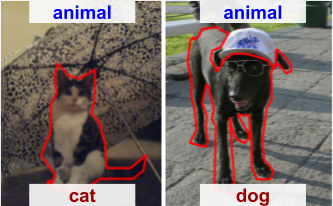}}
    \caption{\color{Blue} parent \color{black} / \color{BrickRed} child}
    \end{subfigure}
    \begin{subfigure}[b]{0.475\linewidth}
    \fbox{\includegraphics[height=2cm]{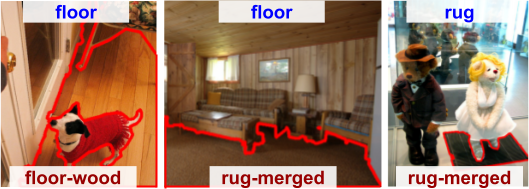}}
    \caption{overlap}
    \end{subfigure}
    \vspace{-5mm}
    \caption{
    Examples of relations: 
    (a) \emph{identity}: both bicycle labels contain similar instances; 
    (b) \emph{parent/child}: the \ade{animal} class contains instances which are either \coco{cat} or \coco{dog}; 
    (c) \ade{floor} and \coco{rug-merged} \emph{overlap} in the middle instance. But each label contains instances which are incompatible with the other label. %
    \vspace{-3mm}
    }
    
    \label{fig:relation_examples}
\end{figure}

Progress in computer vision is fueled by the availability of many different datasets, covering a wide spectrum of appearance domains and annotated for various task types, like ImageNet for classification~\cite{deng09cvpr}, Open Images for detection~\cite{kuznetsova20ijcv}, and KITTI for semantic segmentation of driving scenes~\cite{geiger13ijrr}.
Each of these datasets has its own set of class labels, its own visual definition for each class, its own set of images following a specific distribution, its own annotation protocols, and was labeled by a different group of humans annotators.
As a result, the visual-semantical meaning of a certain label in a particular dataset is unique~\cite{ponce06datasetissues,Torralba11}.
A few examples:
(1) a \ade{sofa} in ADE20k refers to the same visual concept as a \coco{couch} in COCO, even though their class label is different;
(2) ADE20k distinguishes \ade{stool}, \ade{armchair}, and \ade{swivel chair} whereas COCO has a single concept \coco{chair}. Moreover it is unclear if instances of \ade{stool} would adhere to the annotation definition of the \coco{chair} class in COCO;
(3) ADE20k has the labels \ade{floor} and \ade{rug} whereas COCO distinguishes \coco{floor-wood} and \coco{rug-merged}.
These are two ways of categorizing the visual world which are not fully compatible: a full-floor carpet is both a \ade{floor} and a \coco{rug-merged}, while a wooden floor is only a \ade{floor} and a doormat is only a \coco{rug-merged} (see also Fig.~\ref{fig:relation_examples}c).

In this paper we want to automatically discover relations between labels across datasets. We aim to determine if the ADE20k \ade{lake} and COCO \coco{water-other} labels are related in their visual semantics. 
More specifically, are there visual instances which can be described by both labels? And what is the \emph{type} of their relation? Do they represent the same visual concept? Are they in a parent/child relation? Or do they overlap like \ade{floor} and \coco{rug-merged}?
Establishing such relations would enable combining datasets. This is useful for training on larger dataset with more visual concepts and more samples per concept, and also for evaluation purposes.

Joining datasets cannot be done by simply looking at the class label names; how labels across datasets are related really depends on the idiosyncrasies of each dataset involved. Indeed, Lambert et al.~\cite{lambert20cvpr} recently proposed to unify multiple datasets into a single and consistent label space. This required a tremendous amount of manual work:
matching all labels, visually verifying whether labels actually point to the same visual concepts, and re-annotating significant portions of each dataset into a single, mutually exclusive label space. Essentially~\cite{lambert20cvpr} manually solved some of the kind of problems we want to address automatically.
But also their result is prone to similar issues as described, their result depends on choices made for the definitions of labels, the annotation protocol, \emph{etc}. Moreover as the number of datasets continues to grow, such heroic manual joining operation becomes infeasible and it will be necessary to do this automatically.

In this paper, we present methods for the automatic discovery of relations between labels across dataset. We distinguish different relation types (\autoref{tab:relations}):
\emph{identity} (e.g. ADE20k \ade{bicycle} and COCO \coco{bicycle}), \emph{parent/child} (e.g. ADE20k \ade{animal} and COCO \coco{dog}), and \emph{overlap} (e.g. ADE20k \ade{floor} and COCO \coco{rug-merged}). 
We introduce methods to establish these relations by leveraging language cues, visual cues, and a combination of both.

In short, this paper presents an exploration into the discovery of how labels across datasets relate to each other.
Our contributions are as follows:
(1) We introduce a variety of methods to discover the existence of relations between labels across datasets, as well as their type (Sec. \ref{sec:automatic_relations}). These methods include vision, language, and their combination.
(2) We demonstrate that we can effectively and automatically discover label relations between three semantic segmentation datasets: COCO~\cite{caesar18cvpr,lin14eccv,kirillov19cvpr}, ADE20k~\cite{zhou17cvpr}, and Berkeley Deep Drive (BDD)~\cite{fisher20cvpr} (Sec.~\ref{sec:results}).
To evaluate this quantitatively we leverage the MSeg annotations~\cite{lambert20cvpr} to establish ground-truth label relations between these datasets.
Additionally, we show that we can discover relations between different \emph{types} of datasets by applying our method to ILSVRC12 image classification and COCO segmentation (Appendix~\ref{sec:ilsvrc_coco}).
(3) We demonstrate the usefulness of our method in four applications:
\emph{Understand label relations} (Sec.~\ref{sec:understanding_relations}), in which we gain a deeper understanding of what types of relations exist and why they arise in practice;
\emph{Identify missing aspects} (Sec.~\ref{sec:missing_aspects}), where we determine how datasets vary in covering appearance variability of a class;
\emph{Increase label specificity} (Sec.~\ref{sec:label_specificity}), where we can relabel instances of a class at a finer-grained level;
\emph{Predict transfer learning gains} (Sec.~\ref{sec:transfer_learning}), where our label relations can predict the gains brought by transfer learning.

\section{Related Work}\label{sec:related_work}

\para{Dataset creation and evolution.}
In computer vision there is a long standing history to create datasets for training and benchmarking methods.
There are too many to recall here, but interestingly many popular dataset have evolved over time, either by growing the number of images, like ImageNet~\cite{deng09cvpr}, the number of classes, like  PASCAL-VOC~\cite{everingham15ijcv} from 4 classes in 2005 to 20 in 2007, or in the types of annotation, like COCO~\cite{lin14eccv} to COCO-stuff~\cite{cocostuff-dataset} to COCO-panoptic~\cite{kirillov18coco}. 
Other datasets evolve by merging, for example the SUNRGB-D~\cite{zhou14nips} dataset combined imagery from among others NYU-depth-v2~\cite{silberman12eccv} and SUN3D~\cite{xiao13iccv}, while the ADE20K~\cite{zhou17cvpr} dataset contains imagery from SUN~\cite{xiao10cvpr} and Places~\cite{zhou14nips}.

In this paper we use the COCO-panoptic dataset~\cite{caesar18cvpr,kirillov18coco,lin14eccv}, ADE20K~\cite{zhou17cvpr} and BDD~\cite{yu20cvpr}.
Instead of considering these dataset individually, 
we explore how the visual concepts in these dataset \emph{relate} to each other. The relations we find could be used when aiming to combine these datasets or when aiming to train more generic models across different datasets.

\para{Learning over diverse image domains.}
Any single dataset has issues by its design~\cite{ponce06datasetissues}, bias~\cite{Torralba11}, or evaluation robustness~\cite{zendel17cvpr}. Therefore a recent trend is to train or evaluate algorithms over multiple datasets.
For example in the Robust Vision Challenge~\cite{robustvisionchallenge}
participants are asked to evaluate a single trained model over multiple datasets and the winner is based on the average performance. 
To facilitate this, collection of datasets have been introduced, for example, Visual Decathlon for image classification~\cite{rebuffi17nips}, Meta-Dataset for few-shot learning~\cite{triantafillou20iclr}, and MSeg for semantic segmentation~\cite{lambert20cvpr}.

Training tactics to successfully use multiple datasets differ, 
from training a single model with different heads over all datasets jointly~\cite{kokkinos17cvpr}, 
to learn in stages, \emph{i.e.}, first on ImageNet, then tune on COCO and finally fine-tune on PASCAL-VOC~\cite{mensink21pami};
and from using manually merged labels~\cite{bevandic22wacv,lambert20cvpr},
to post-hoc merging of labels for detection~\cite{zhou22cvpr}.
In contrast to these approaches, our aim is not to train a new model with better classifiers, but we aim to analyze more fundamentally how datasets relate to each other.

\para{Zero-shot and open set segmentation.}
For both zero-shot and open set segmentation the goal is to obtain pixel-wise predictions for never-seen labels using zero training examples~\cite{bucher19neurips,ghiasi21arxiv}. 
Both aim to learn classifiers which generalize the set of training classes to a fixed set of never seen labels~\cite{bucher19neurips} or open vocabulary queries~\cite{ghiasi21arxiv}. 
This works by establishing (language) based relations between seen and unseen classes, for example based on large scale contrastive pre-training on images and textual queries~\cite{jia21icml_align,radford21icml_clip}.
In contrast to these methods, our aim is not to train generalizable classifiers, but to find the relations between visual concepts in both datasets, for which we can make use of the available annotations.
\section{Method}\label{sec:automatic_relations}

In this paper we want to automatically discover relations between class labels across two given datasets $A$ and $B$.
We consider all possible pairs $\langle a, b\rangle$ of labels $a$ in $A$ and $b$ in $B$. 
For each pair we want to determine if they are \emph{related}, i.e. where there are visual instances which are covered by both the definition of $a$ and $b$,
and we also want to determine the \emph{type} of the relation (Tab.~\ref{tab:relations}). 

\begin{table}[t]
\centering
\begin{tabular}{m{97.5mm}c}
\toprule
{\bf Identity} \quad Label $a$ in one dataset indicates the same visual concept as label $b$ in another dataset. For example \ade{sofa} in ADE20k and \coco{couch} in COCO represent the same visual concept.
& \includegraphics[valign=m,height=6mm]{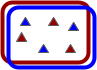}\\
{\bf Parent/child} \quad A subcategory relationship. For example, \ade{animal} in ADE20k is the parent of \coco{cow} in COCO. 
& \includegraphics[valign=m,height=6mm]{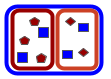}\\
{\bf Overlap} \quad 
Label $a$ in one dataset and label $b$ in another describe visual concepts which are not the same even though their sets of instances intersect. 
For example, the ADE20k \ade{floor} and COCO \coco{rug-merged} both describe a floor-covering carpet. Yet both concepts are broader in a mutually exclusive way: \ade{floor} also includes a wooden floor which is not a \coco{rug-merged}. Conversely, \coco{rug-merged} also includes a rug which can be picked up which is not a \ade{floor}.
 & \includegraphics[valign=m,height=6mm]{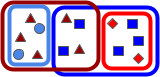} \\
{\bf Part-of} \quad Label $a$ in one dataset captures parts of instances of label $b$ in another dataset. For example, \coco{roof} in COCO describes part of an instance of \ade{house} in ADE20k. 
& \includegraphics[valign=m,height=6mm]{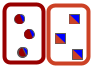} \\
\bottomrule
\end{tabular}
\vspace{-0mm}
\caption{Definition of types of label relations we distinguish. In this paper we aim to automatically identify all relations (Sec.~\ref{sec:automatic_relations}) except the \emph{part-of} relation. 
}
\label{tab:relations}
\end{table}

We distinguish \emph{identity}, \emph{parent/child}, \emph{overlap}, and \emph{part-of} (focusing mostly on the first three).
Importantly, the existence of a relation between two labels and its type cannot be derived simply by considering their names. 
Instead, they are specific to the pair datasets from which they originate, because they depend on the design and construction of each dataset.

\subsection{Discovering relations using visual information}\label{sec:visual_relations}

\begin{figure}[t]
    \centering
    \begin{subfigure}[b]{.95\textwidth}
    \includegraphics[width=\textwidth]{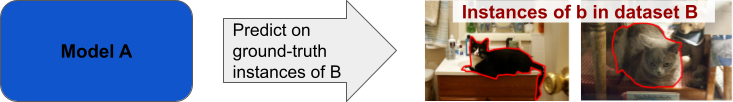}
    \vspace{2mm}
    \end{subfigure}
    \begin{subfigure}[b]{.95\textwidth}
    \includegraphics[width=\textwidth]{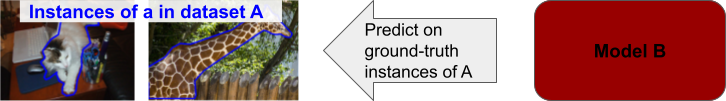}
    \end{subfigure}    
    \caption{
    Illustration of how to obtain label link scores between \Ade{animal} and \Coco{cat}, we estimate $S_{a \rightarrow b}(\ade{animal}, \coco{cat})$ using the model trained on ADE20K and $S_{b \rightarrow a}(\coco{cat}, \ade{animal})$ using the model trained on COCO.}
    \label{fig:main_idea}
\end{figure}

We first discuss how we discover relations and their type using purely visual information (Fig.~\ref{fig:main_idea}). 
We do this in the context of semantic segmentation, but our method would also work for object detection and we apply it to discover relations between ILSVRC classification and COCO segmentation in Appendix~\ref{sec:ilsvrc_coco}.

To determine whether there exist a relation between label $a$ in dataset $A$ and $b$ in dataset $B$, we use annotated 
{\em instances}\footnote{An instance is either a single object (for thing classes, e.g. \class{cat}, \class{car}), or the union of all regions of a stuff class (e.g. \class{grass}, \class{water}), following the panoptic definition~\cite{kirillov19cvpr}.} of these classes in their respective datasets.
We use a model $p_A$ trained on dataset A to obtain predictions $p_A(a|i_b)$ for label $a$ for an instance $i_b$ with label $b$ from dataset B.
Next, we average these predictions: 
\begin{equation}
    S_{a \rightarrow b} = \frac{1}{n_b} \sum_{i_b \in B} p_A(a|i_b)    
\end{equation}
where $n_b$ is the number of instances of label $b$ in dataset $B$.
Intuitively, this measures how likely it is that the instances of $i_{b=\coco{cat}}$ from the COCO dataset ($B$) would be called $a=\ade{animal}$ according to the model trained on ADE20k (dataset $A$).
Similarly we obtain $S_{b \rightarrow a}$ by aggregating predictions of $p_B$ over instances of dataset $A$.
The final score is the average: $R_{a, b} = (S_{a \rightarrow b} + S_{b \rightarrow a}) / 2$.
To determine whether there is a relation between label $a$ and label $b$, we simply threshold $R_{a, b}$.
This results in a set $\mathcal{R}$ of binary relations.

Experimentally we evaluate two different prediction models $p_A(a|i_b)$:
\begin{itemize}
    \item \emph{Pixel Probabilities}: applying a segmentation model trained on dataset $A$ directly on instances of dataset $B$. To convert to instance probabilities we average the pixel-wise probabilities over all pixels of the instance;
    \item \emph{Visual Embeddings}: we extract instance visual features for both dataset $A$ and dataset $B$ by aggregating the pixel-wise visual features, using the same segmentation model (trained on dataset $A$) without the classification head. Then we use a 1-Nearest Neighbour classifier. This results in a binary prediction \emph{i.e.} $p_A(a|i_b)$ is either 1 or 0. We do the analogue for $p_B(b|i_a)$.
\end{itemize}

\para{Training details.} %
We train semantic segmentation models using an HRNetV2-W48~\cite{wang20pami} backbone with a linear pixel-wise prediction head and a softmax-loss. This results in a strong model for semantic segmentation~\cite{lambert20cvpr,mensink21pami,wang20pami}.
We unify the training setup to make the models compatible across datasets, using color normalization, horizontal flipping, random crop and resize to $713\times713$. 
We optimize using SGD with momentum, with $\textrm{lr}=0.01$ decreased by a factor 10 after $2/3$rd of the number of training steps (optimized per dataset). 

While for semantic segmentation typically the \class{background} class is ignored during training and evaluation we find it useful to incorporate it explicitly.
The \class{background} prediction can be interpreted as the model predicting \emph{none of the classes from my label space}.
Moreover, we find it beneficial to only aggregate over \emph{easy} instances to factor out errors introduced by miss-classification of difficult instances.
To do so we use instances which are classified correctly by the model trained on the same dataset.
More specifically, we define instances to be easy for the pixel probability method if $p_B(b|i_b)>0.5$. They are easy for the visual embedding method if $p_B(b|i_b) = 1$.

\subsection{Relation type discovery}
\label{sec:relation-type-disco}
We estimate the \emph{type} of relation (Tab.~\ref{tab:relations}) in two different ways, one based on set theory and the other on the degree of asymmetry between $S_{a \rightarrow b}$ and $S_{b \rightarrow a}$.

\para{\emph{Set theory}.}
To derive the relation types we make two assumptions:
(1) There is only a relation between label $a$ and label $b$ if there are instances which can be categorized as both $a$ and $b$, so $\langle a,b\rangle \in \mathcal{R}$;
(2) Labels from the same dataset are mutually exclusive.
Then we derive the types between $a_k$ and $b_l$ as follows: 
\begin{itemize}
    \item \emph{identity}:
    $a_k$ and $b_l$ have an identity relation when neither $a_k$ nor $b_l$ has a relation with another label.
    More formally,
    $\langle a_k, b_l\rangle \in \mathcal{R}$, but $\nexists a_m, \langle a_m, b_l\rangle \in \mathcal{R}, a_m \not= a_k$ and 
    $\nexists b_n, \langle a_k, b_n\rangle \in \mathcal{R}, b_n \neq b_l$.
    
    \item \emph{parent/child}: 
    A label $a_k$ is a parent if it is related to at least two labels in $B$ (including $b_l$), which are not related to any other label in $A$.
    More formally, 
    for at least two labels $b_l$ and $b_n$, $b_l \neq b_n$, it holds that 
    $\langle a_k, b_l\rangle \in \mathcal{R}$ and 
    $\langle a_k, b_n\rangle \in \mathcal{R}$. 
    Yet, 
    $\nexists a_m, \left[ \langle a_m, b_l\rangle \in \mathcal{R} \lor \langle a_m, b_n\rangle \in \mathcal{R} \right], a_m \neq a_k$.
    Analogously, $a_k$ is a child of $b_l$ if their roles are reversed.
    
    \item \emph{overlap}:
    both labels $a_k$ and $b_l$ are used in multiple relations. Formally,
    $\langle a_k, b_l\rangle \in \mathcal{R}$ and 
    $\exists a_m, \langle a_m, b_l\rangle \in \mathcal{R}, a_m \neq a_k$ and
    $\exists b_n, \langle a_k, b_n\rangle \in \mathcal{R}, b_n \neq b_l$.
\end{itemize}

\para{\emph{Score Asymmetry}.}
We exploit the asymmetry between $S_{a \rightarrow b}$ and $S_{b \rightarrow a}$ to provide the type of the relation.
Intuitively, for a \emph{parent-child} relation, we expect an \ade{animal} classifier to give high scores on \coco{cat} instances, while 
the \coco{cat} classifier only gives high scores on \emph{some} of the \ade{animal} instances. 
Therefore, a large asymmetry between $S_{a \rightarrow b}$ and $S_{b \rightarrow a}$ suggests that the labels are in a \emph{parent-child} relation. 
Given a pair of labels $(a, b) \in \mathcal{R}$ we derive the label as follows:
1) $a$ is a \emph{parent} of $b$, if $\frac{S_{a \rightarrow b}}{S_{b \rightarrow a}} > T$; else
2) $a$ is a \emph{child} of $b$, if $\frac{S_{b \rightarrow a}}{S_{a \rightarrow b}} > T$; otherwise
3) $a$ and $b$ are in an \emph{identity} relation. 
Note this method cannot predict \emph{overlap}.

\subsection{Predicting relation types using language}
\label{sec:language-relations}

We introduce two baseline methods which use language to discover relations.

\para{\emph{WordNet.}}
We use the WordNet~\cite{miller95acm} taxonomy and its graphical structure.
Specifically, we map each class label to a WordNet noun-synset.
Then, if $a$ and $b$ map to the same synset, they are in an \emph{identity} relation.
When the synset of $a$ is an ancestor of the synset of $b$, then $a$ is a \emph{parent} of $b$.
If two synsets share at least one descendant, they are in an \emph{overlap} relation. For example, in WordNet \class{car} and \class{truck} overlap since they both have \class{minivan} as a descendent.

For each pair of labels ($a$,$b$) we estimate the \texttt{path similarity} between the two synsets, which is based on the proximity of their nearest common ancestor.
Then we add 1 if $a$ and $b$ have a relation according to the taxonomy. This yields a dense matrix $R$, with pairs discovered as \emph{identity} have a strength of 2, as \emph{parent}, \emph{child}, or \emph{overlap} have a strength between 1 and 2, and the rest between 0 and 1.

\para{\emph{Word2Vec.}}
Our second baseline uses Word2Vec~\cite{mikolov13iclr}, based on the publicly available model trained on Wikipedia~\cite{word2vectfhub}. 
This maps each word to a 500-D embedding vector. The score between each paper of labels $a,b$ is based on the cosine similarity between their embeddings.
Since this is a symmetric similarity, we can only use the \emph{set theory} method to determine relation types.

\subsection{Discovering relations by combining vision and language}
\label{sec:combo-disco}

We combine our Visual Embeddings method with our WordNet method. We multiply the strength of the visual relation $R_{a, b}$ by a constant factor $n$ if the synset of $a$ and the synset of $b$ are related according to the taxonomy (i.e. we discover \emph{identity}, \emph{parent}, \emph{child}, or \emph{overlap}).

To discover the relation type, we combine the visual \emph{asymmetry} method and the WordNet predictions: If according to WordNet $a$ and $b$ are in an \emph{identity} relation, we enlarge threshold $T$ of the \emph{asymmetry} method by a factor $m$. This makes it more likely that \emph{identity} will be predicted. Similarly, when according to WordNet $a$ and $b$ are in a \emph{parent}/\emph{child} relation, we reduce $T$ by a factor $m$.
\subsection{Evaluation}\label{sec:evaluation}

\begin{figure}[t]
    \centering
    \vspace{-.3cm}
    \includegraphics[width=.8\textwidth]{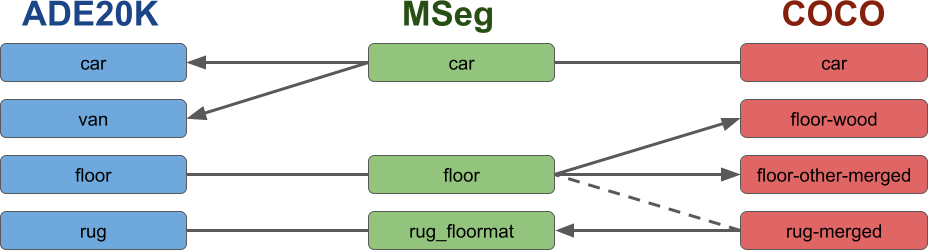}
	\caption{To create relations between ADE20k and COCO, we first establish relations between each individual dataset and MSeg. Based on set theory (Sec.~\ref{sec:visual_relations}) we establish \emph{identity} (\mytikzline), \emph{parent}/\emph{child} (\mytikzarrow), and \emph{overlap} (\mytikzdashedline) relations. Afterwards, through MSeg we derive direct relations between ADE20k and COCO.} 
	\label{fig:gt_creation}
	\vspace{-0.2cm}
\end{figure}

To evaluate how well we are able to automatically discover relations between labels across datasets, we first establish ground-truth relations\footnote{\scriptsize Available at: \url{https://github.com/google-research/google-research/tree/master/missing_link}}.
We leverage the MSeg dataset~\cite{lambert20cvpr}, who manually constructed a unified label space across a variety of different datasets, 
which we refer to as {\em MSeg labels} (Fig.~\ref{fig:gt_creation}).
Based on this, we first map dataset $A$ and dataset $B$ to the MSeg label space, and then create direct relations between labels in $A$ and $B$.

\para{Establishing relations to the MSeg label space.}
The MSeg dataset provides for all dataset which they cover a new ground-truth in the unified MSeg label space. This MSeg ground-truth covers a different set of labels than the original ground-truth for each dataset; for each dataset the authors merged some classes and re-annotated other classes to obtain a consistent labeling of each dataset according to the MSeg label space~\cite{lambert20cvpr}. We compare the MSeg ground-truth with the original ground-truth to establish relations.

In particular, we count how many times an instance with a particular label in the original label space is relabeled to each MSeg label. For an instance to count being relabeled, more than 50\% of its pixels need to have been relabeled. This makes the process robust against small manual corrections made during the MSeg relabeling effort.
We manually inspect all label pairs with a positive count, and remove them when this is caused by a human error (low counts typically help identify these cases).
For example, a few instances of COCO \coco{tent} have been relabeled to the MSeg \mseg{kite}, while these labels are clearly unrelated. %
All remaining pairs are considered as related in our ground-truth.

To derive the type of relation, we apply the \emph{set theory} method from Sec.~\ref{sec:visual_relations},
 and then manually investigate all relations. We found that almost no human correction was needed at this stage. The only exception was that a few relations were changed to \emph{part-of}, which the set theory method cannot automatically produce. For example, COCO \coco{roof} is \emph{part-of} the MSeg \mseg{building}.

\para{Establishing relations between A and B.}
Through the MSeg labels, we can directly relate the original labels between datasets (Fig.~\ref{fig:gt_creation}). The type of relation depends on the type of the two individual relations with MSeg. When both relations are \emph{identity}, the resulting relation is that as well. Two consecutive \emph{child} relations or one \emph{identity} and one \emph{child} relation result in a \emph{child} relation. For example, ADE20k \ade{van} is a \emph{child} of \Coco{car}. The \emph{parent} relation is analogous to \emph{child}. 
If one relation is \emph{part-of}, the resulting relation is \emph{part-of} as well. For all other cases, we manually inspect visual examples to determine the relation type. Often this happens for \emph{overlap} relations. But for example both COCO \coco{person} and the ADE20k \ade{person} have been sub-categorized by MSeg in \mseg{person}, \mseg{bicyclist}, \mseg{motorcyclist}, and \mseg{rider\_other}. It requires manual inspection to verify that both \class{person} labels represent the same concept and hence have an \emph{identity} relation.
As before, after these steps we perform a final quality control by manually inspecting visual examples of label pairs.

\para{Quantitative evaluation.} For two datasets, we compare our automatically predicted relations with the ground-truth we just established. We evaluate how good our methods are in predicting whether any relation is present, regardless of its type. 
To do so we order all possible label pairs according to their predicted strength, and calculate a Precision-Recall (PR) curve and its associated Area Under the Curve (AUC).
We also measure how well our methods predicts relation types, where we also consider \emph{no relation} predictions. We measure accuracy for each predicted type and average them to obtain an overall accuracy.

\begin{figure}[t]
    \centering
    \vspace{-.4cm}
    \includegraphics[width=0.32\linewidth]{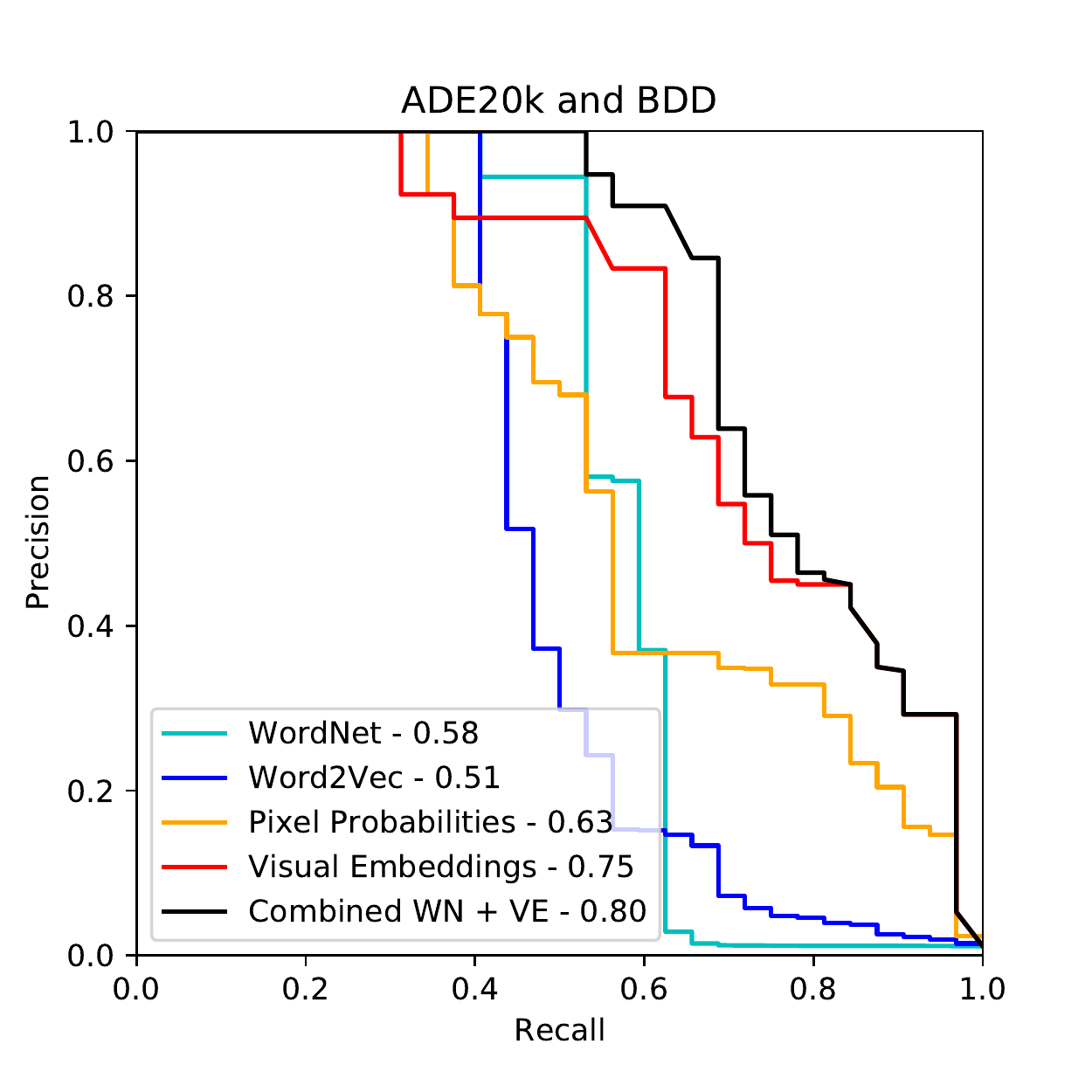}
    \includegraphics[width=0.32\linewidth]{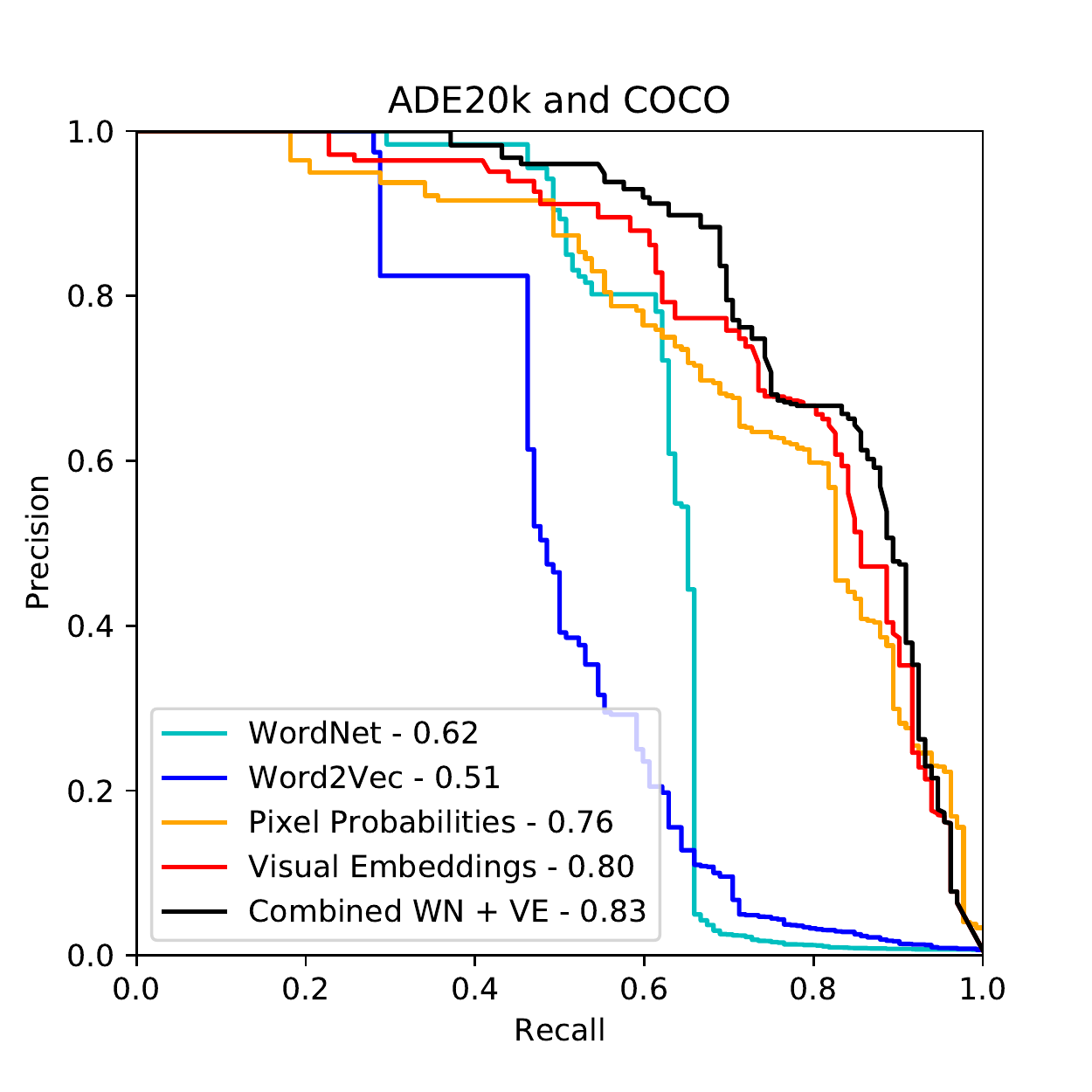}
    \includegraphics[width=0.32\linewidth]{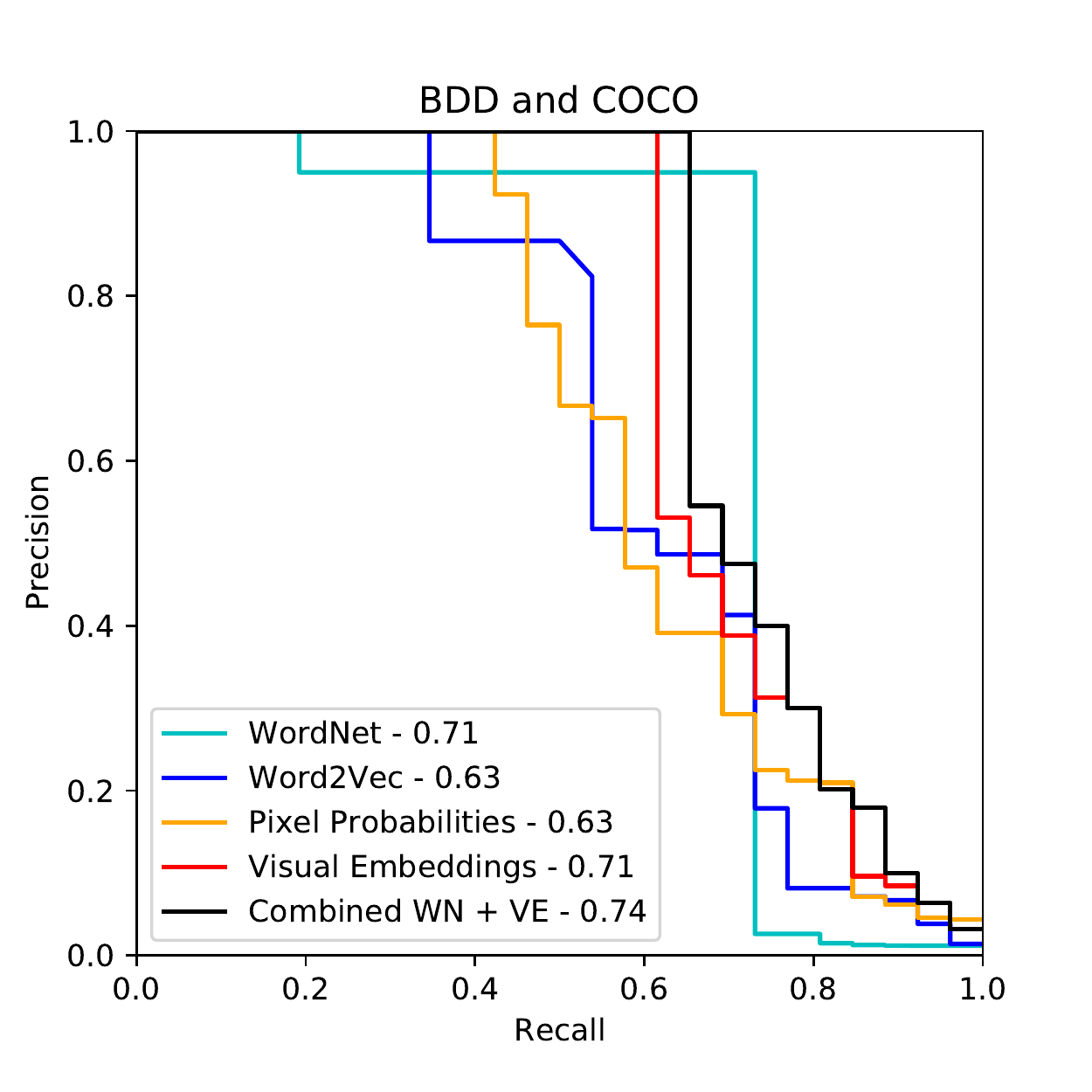}
    \vspace{-5mm}
    \caption{Precision recall curves for different methods for (binary) label relation predictions. The visual methods perform (much) better than language-only methods and combining vision and language gives best performance.\vspace{-3mm}}
    \label{fig:pr_curve}
\end{figure}

\section{Results}\label{sec:results}

As our main experiment we apply and evaluate our method on all three possible pairs of the following semantic segmentation datsets:
(1) ADE20k~\cite{zhou17cvpr}, a dataset of consumer photos, where we consider the 150 most frequent class labels as is common practice;
(2) COCO Panoptic~\cite{caesar18cvpr,lin14eccv,kirillov19cvpr}, which also contains consumer photos, with 133 classes;
(3) Berkeley Deep Drive~\cite{yu20cvpr}, a driving dataset containing 19 classes.
For ease of exposition, we write the names of class labels for ADE20k in \ade{blue}, for COCO in \coco{red}, and for BDD in \bdd{violet}.

Additionally, in Appendix~\ref{sec:ilsvrc_coco} we demonstrate that we can establish relations between labels of different types of datasets by applying our method to ILSVRC classification and COCO segmentation.

\para{Relation discovery.}
The Precision-Recall curves in Fig.~\ref{fig:pr_curve} show that the language-based models are generally outperformed by the vision-based models. 
The model based on WordNet works better than Word2Vec, because Word2Vec gives high scores for labels which are semantically related but do not refer to the same object. For example, the Word2Vec cosine similarity between \ade{shower} and \coco{toilet} is 0.72 while these classes are really disjoint. The WordNet-based method has high accuracy for labels in an \emph{identity} or \emph{parent/child} relation according to the taxonomy, but a low recall for many other relations.
Among the vision models, the Visual Embeddings method consistently outperforms the Pixel Probability method (Sec. \ref{sec:visual_relations}).
Finally, we obtain the best performance when combining WordNet with Visual Embeddings.

\para{Relation type classification.}
Before we can determine relation types, we note that the \emph{Set Theory} and \emph{Score Asymmetry} methods have thresholds (Sec. \ref{sec:relation-type-disco}). We establish these by optimizing accuracy with respect to the predictions made by the WordNet \emph{taxonomy}. While the WordNet \emph{taxonomy} method may not be fully accurate, as long as it is an unbiased estimate its optimal thresholds will also hold for the real ground-truth - which is indeed what we found.

The results in Tab.~\ref{tab:type_accuracy} generally align with our previous observations: WordNet is the best language-based model but the vision-based models work even better. Again, the Visual Embeddings method outperforms all others. From the two ways to determine the relation type, the one based on \emph{Score Asymmetry} works best.
Intuitively, it makes sense that this is a powerful mechanism: we expect an \class{animal} model to always yield high scores on \class{cat} instances, whereas a \class{cat} model will not give high scores to all \class{animal} instances. As before, the combination of WordNet and Visual Embeddings gives the best results.

\begin{figure}[t]
    \centering
    \begin{subfigure}[b]{0.3\linewidth}
    \includegraphics[width=\linewidth]{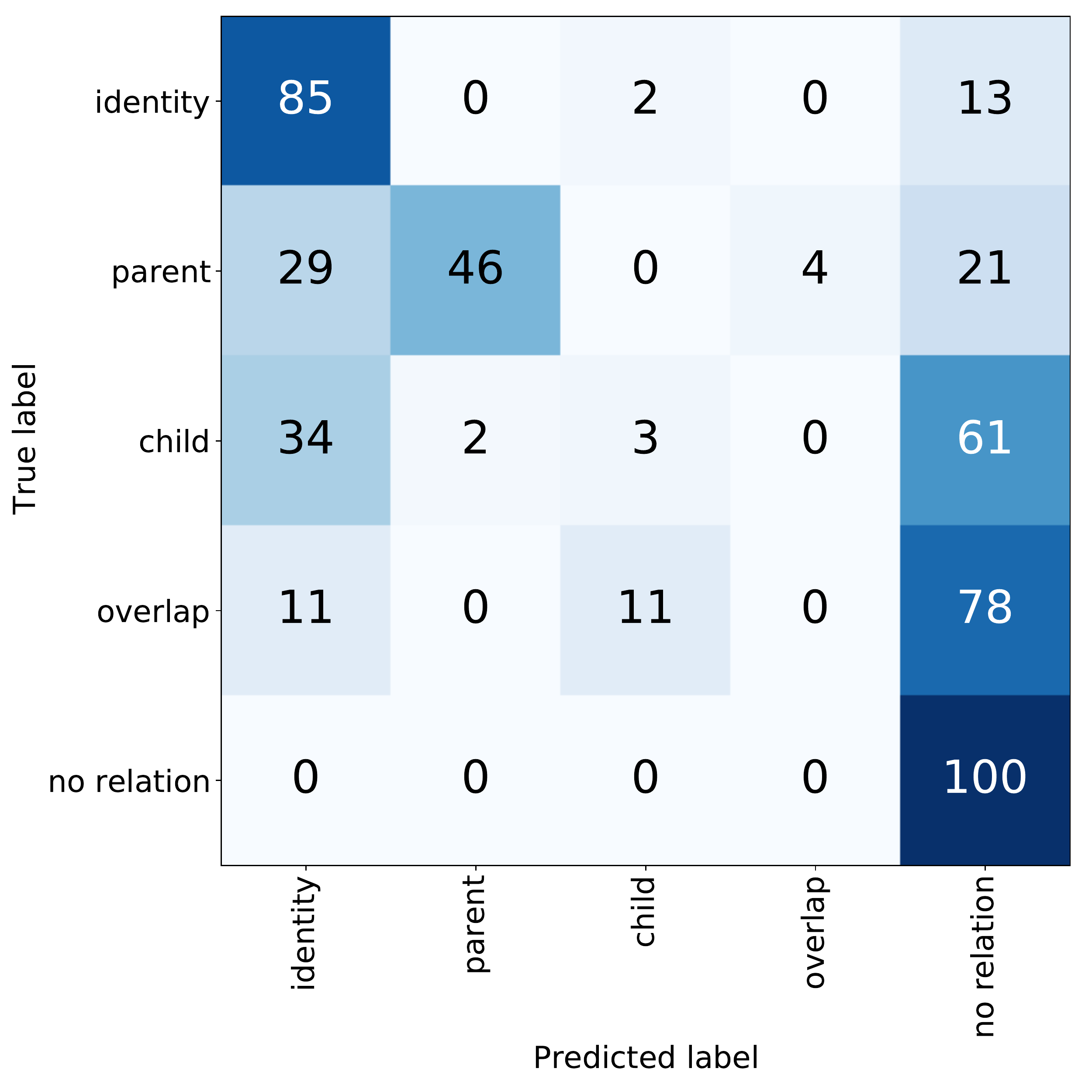}
    \end{subfigure}
    \hfill
    \begin{subfigure}[b]{0.3\linewidth}
    \includegraphics[width=\linewidth]{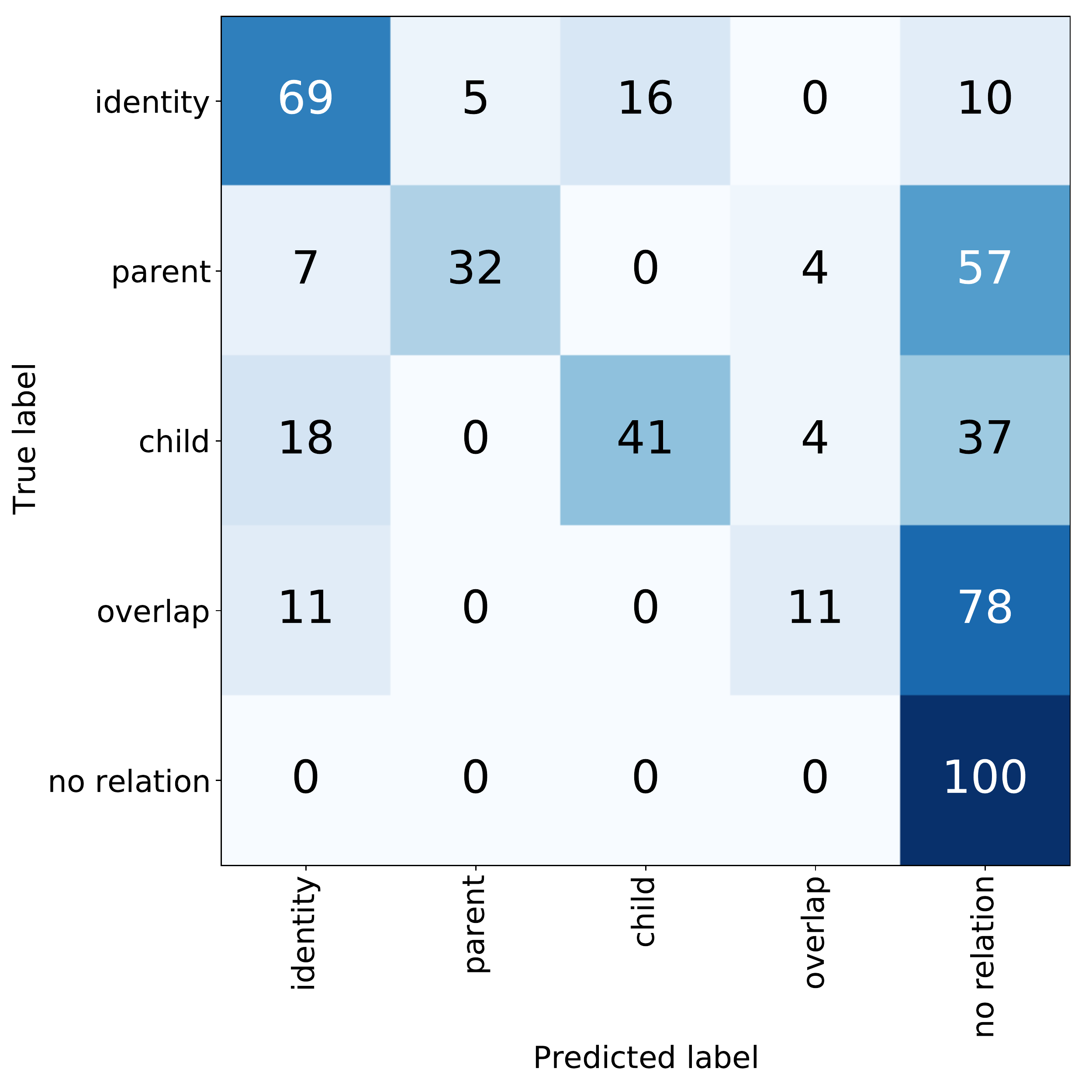}
    \end{subfigure}
    \hfill
    \begin{subfigure}[b]{0.3\linewidth}
    \includegraphics[width=\linewidth]{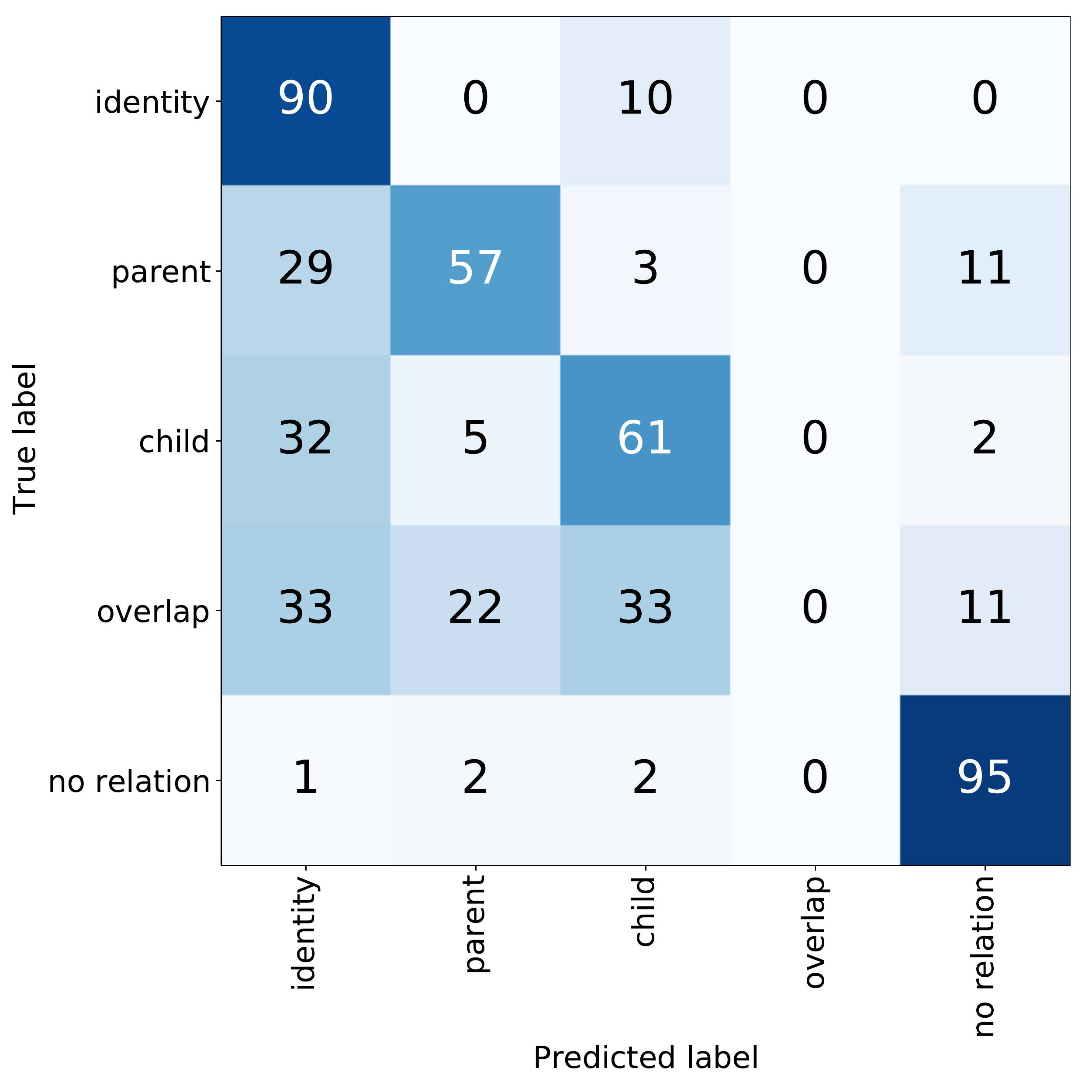}
    \end{subfigure}
    \caption{Confusion matrices for relation types between ADE20K and COCO using WordNet-taxonomy (\emph{left}), Embeddings with Set Theory (\emph{middle}), and Embedding with Asymmetry (\emph{right}).}
    \label{fig:confusion}
\end{figure}

In Fig.~\ref{fig:confusion} 
we show the full confusion matrices for relations between ADE20k and COCO discovered by the WordNet \emph{taxonomy}, Visual Embeddings with  \emph{Set Theory}, and Embeddings with \emph{Asymmetry} methods. We can see that the WordNet taxonomy predicts both the identity and `no relation' pretty well, but tends to over-predict `no relation'.
The Embeddings with \emph{Set Theory} also over-predicts `no relation', it is slightly worse in `identity' but better in `child' and `overlap'. Embeddings with \emph{Asymmetry} is significantly better in parent and child relations. However, it cannot predict `overlap'.

\begin{table}[t]
\centering
\resizebox{\textwidth}{!}{
\begin{tabular}{lc@{\hspace{1.5mm}}@{\hspace{1.5mm}}c@{\hspace{2mm}}@{\hspace{2mm}}cc@{\hspace{1.5mm}}@{\hspace{1.5mm}}cc@{\hspace{4mm}}@{\hspace{2mm}}c}
\toprule
& \multicolumn{2}{c}{Language} & \multicolumn{4}{c}{Vision} & Vision+Language \\
\midrule
& WordNet & Word2Vec & \multicolumn{2}{c}{Pixel Predictions} & \multicolumn{2}{c}{Visual Embeddings} & WordNet+Embeddings \\
& \emph{taxonomy} & \emph{set theory} & \emph{set theory} & \emph{asymmetry} & \emph{set theory} & \emph{asymmetry} & \emph{taxonomy}+\emph{asymmetry} \\
\midrule
ADE20k, BDD & 46 & 37 & 56 & 54 & 56 & 55 & 57 \\
ADE20k, COCO & 47 & 38 & 47 & 60 & 51 & 61 & 62 \\
BDD, COCO & 46 & 38 & 46 & 48 & 49 & 51 & 53 \\
\midrule
{\bf average} & 46 & 38 & 50 & 54 & 52 & 56 & 57 \\
\bottomrule
\end{tabular}
}
\caption{Accuracy (in percentage) of estimating relation types. Our vision-based models outperform language-only models for all pairs of datasets. and combining works best.}
\label{tab:type_accuracy}
\end{table}

\section{Applications}\label{sec:analysis}

We apply our method to four applications: understand label relations (Sec.~\ref{sec:understanding_relations}), identify missing aspects (Sec.~\ref{sec:missing_aspects}), increase label specificity (Sec.~\ref{sec:label_specificity}), and predict transfer learning gains (Sec.~\ref{sec:transfer_learning}).

\subsection{Understand Label Relations}\label{sec:understanding_relations}
\begin{figure}[t!]
    \vspace{-0.3cm}
    \makebox[\linewidth][c]{%
    \begin{overpic}[height=2.5cm, trim={2.9cm, 1.5cm, 2.5cm, 1.4cm}, clip]
    {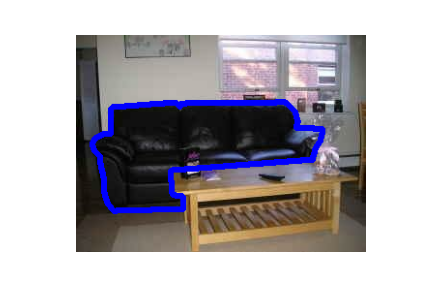}
    \put(40,2){\colorbox{white!80}{\ade{sofa}}}
    \end{overpic}
    \begin{overpic}[height=2.5cm, trim={2.9cm, 1.5cm, 2.45cm, 1.4cm}, clip]
    {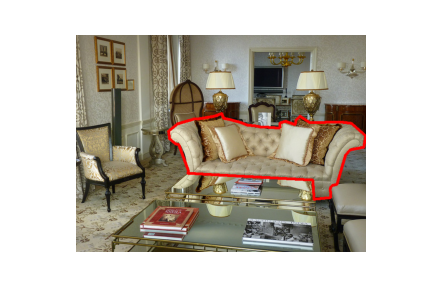}
    \put(40,2){\colorbox{white!80}{\coco{couch}}}
    \end{overpic}
    \hspace{.1cm}
    \begin{overpic}[height=2.5cm, trim={3.2cm, 1.5cm, 5.0cm, 1.4cm}, clip]
    {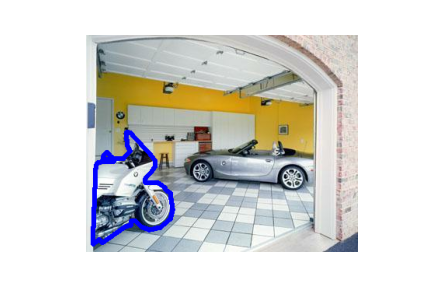}
    \put(25,3){\colorbox{white!80}{\ade{minibike}}}
    \end{overpic}
    \begin{overpic}[height=2.5cm, trim={2.7cm, 1.5cm, 3.3cm, 1.5cm}, clip]
    {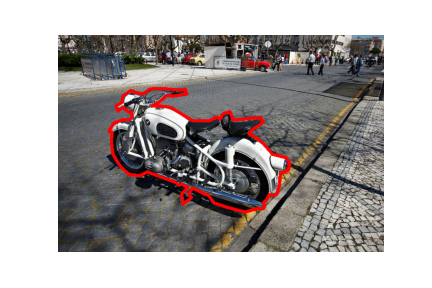}
    \put(25,3){\colorbox{white!80}{\coco{motorcycle}}}
    \end{overpic}
    \hspace{.1cm}
    \begin{overpic}[height=2.5cm, trim={4.8cm, 1.7cm, 4.6cm, 1.4cm}, clip]
    {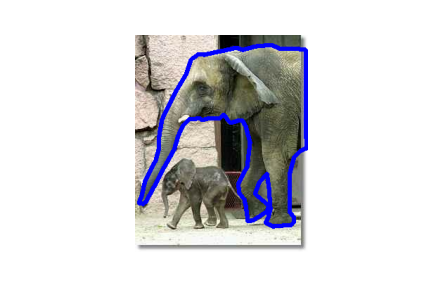}
    \put(20,3){\colorbox{white!80}{\ade{animal}}}
    \end{overpic}
    \begin{overpic}[height=2.5cm, trim={5.3cm, 1.5cm, 5.2cm, 1.4cm}, clip]
    {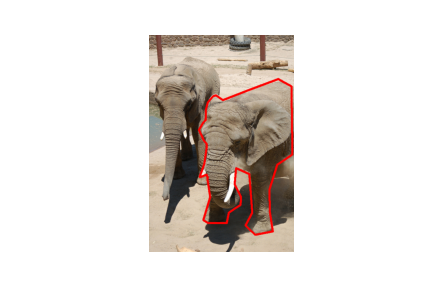}
    \put(2,3){\colorbox{white!80}{\coco{elephant}}}
    \end{overpic}
    } 
    \vspace{-.1cm}
    \begin{tikzpicture}
    \draw[dashed] (-10,0) -- (3,0);
    \end{tikzpicture}
    \vspace{-.1cm}
    
    \makebox[\linewidth][c]{%
    \begin{overpic}[height=2.5cm, trim={3.3cm, 1.5cm, 3.0cm, 1.4cm}]
    {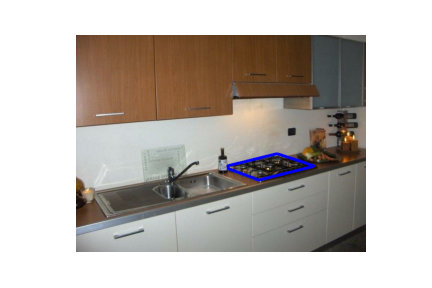}
    \put(30,0){\colorbox{white!80}{\ade{stove}}}
    \end{overpic}
    \begin{overpic}[height=2.5cm, trim={3.3cm, 1.5cm, 3.0cm, 1.4cm}]
    {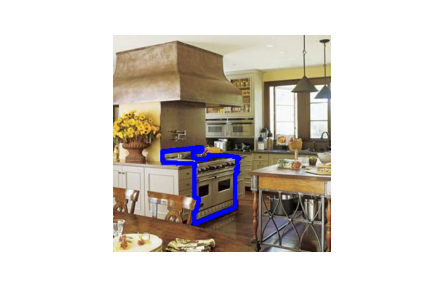}
    \put(30,0){\colorbox{white!80}{\ade{stove}}}
    \end{overpic}
    \begin{overpic}[height=2.5cm, trim={5.2cm, 1.5cm, 3.0cm, 1.2cm}]
    {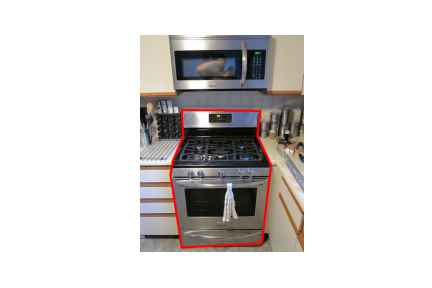}
    \put(20,0){\colorbox{white!80}{\coco{oven}}}
    \end{overpic}
    \begin{overpic}[height=2.5cm, trim={3.3cm, 1.5cm, 3.0cm, 1.4cm}]
    {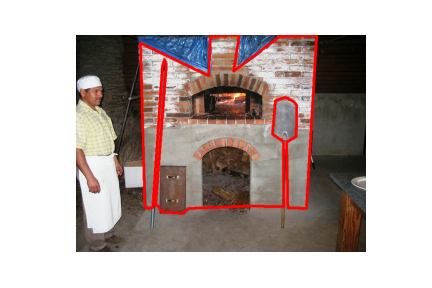}
    \put(40,0){\colorbox{white!80}{\coco{oven}}}
    \end{overpic}
    } \\ 
    \vspace{-.1cm}
    \begin{tikzpicture}
    \draw[dashed] (-10,0) -- (3,0);
    \end{tikzpicture}
    \vspace{-.1cm}
    
    \makebox[\linewidth][c]{%
    \begin{overpic}[height=2.5cm, trim={4.1cm, 1.5cm, 3.7cm, 1.4cm}]
    {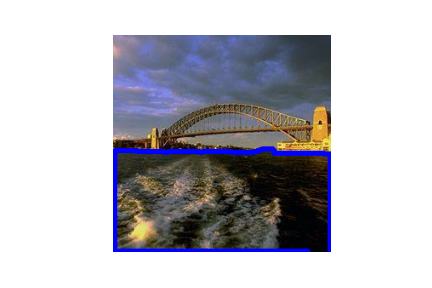}
    \put(30,0){\colorbox{white!80}{\ade{water}}}
    \end{overpic}
    \hspace{.1cm}
    \begin{overpic}[height=2.5cm, trim={3.0cm, 1.5cm, 2.6cm, 1.4cm}]
    {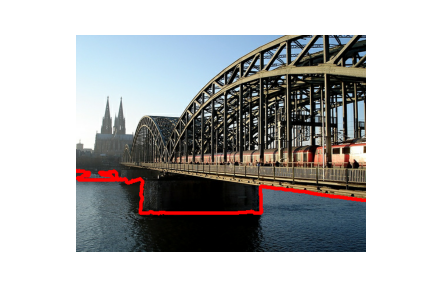}
    \put(35,0){\colorbox{white!80}{\coco{sea}}}
    \end{overpic} 
    \hspace{1.1cm}
    \begin{overpic}[height=2.5cm, trim={2.4cm, 1.5cm, 6.6cm, 1.4cm}, clip]
    {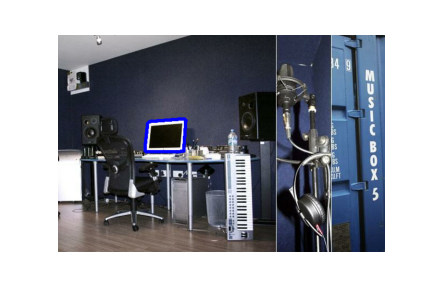}
    \put(5,0){\colorbox{white!80}{\ade{crt screen}}}
    \end{overpic}
    \hspace{.1cm}
    \begin{overpic}[height=2.5cm, trim={3.0cm, 1.5cm, 2.8cm, 1.4cm}, clip]
    {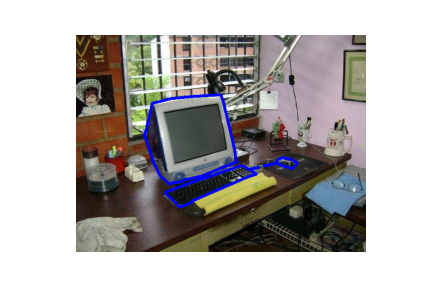}
    \put(30,0){\colorbox{white!80}{\ade{computer}}}
    \end{overpic}
    } \\ 
    \vspace{.2cm}
    
    \makebox[\linewidth][l]{%
    \hspace{-0.5cm}
    \begin{overpic}[height=2.5cm, trim={2.8cm, 1.5cm, 2.4cm, 1.4cm}]
    {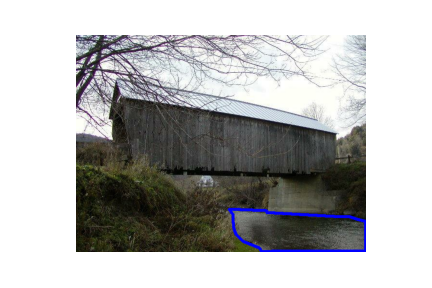}
    \put(10,0){\colorbox{white!80}{\ade{water}}}
    \end{overpic}
    \hspace{.1cm}
    \begin{overpic}[height=2.5cm, trim={4.7cm, 1.5cm, 4.2cm, 1.4cm}]
    {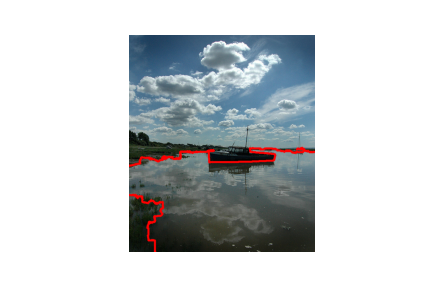}
    \put(30,0){\colorbox{white!80}{\coco{river}}}
    \end{overpic}
    \hspace{1.2cm}
    \begin{overpic}[height=2.5cm, trim={3.0cm, 1.5cm, 3.0cm, 1.4cm}, clip]
    {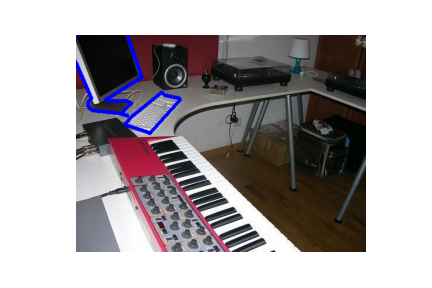}
    \put(30,0){\colorbox{white!80}{\ade{computer}}}
    \end{overpic}
    \hspace{.0cm}
    \begin{overpic}[height=2.5cm, trim={3.0cm, 1.5cm, 3.0cm, 1.4cm}, clip]
    {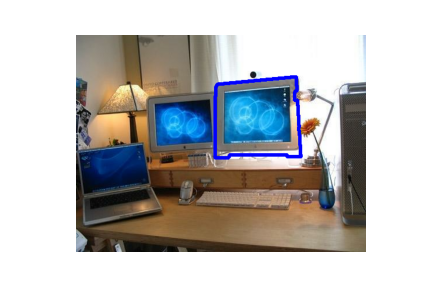}
    \put(30,0){\colorbox{white!80}{\ade{monitor}}}
    \end{overpic}
    } \\ 
    \vspace{.2cm}
    
    \makebox[\linewidth][l]{%
    \hspace{-0.5cm}
    \begin{overpic}[height=2.5cm, trim={4.2cm, 1.5cm, 3.8cm, 1.4cm}]
    {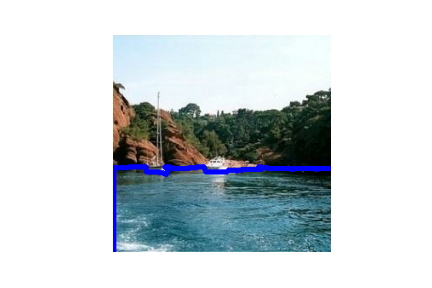}
    \put(35,0){\colorbox{white!80}{\ade{river}}}
    \end{overpic} 
    \hspace{.1cm}
    \begin{overpic}[height=2.5cm, trim={3.3cm, 1.5cm, 3.0cm, 1.4cm}]
    {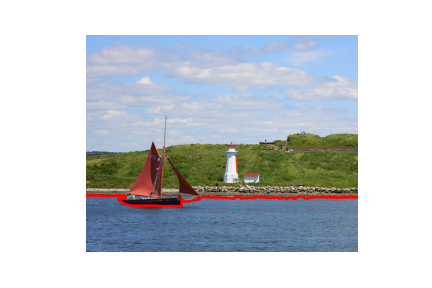}
    \put(35,0){\colorbox{white!80}{\coco{water}}}
    \end{overpic}
    \hspace{1.2cm}
    \begin{overpic}[height=2.5cm, trim={3.3cm, 1.5cm, 3.0cm, 1.4cm}, clip]
    {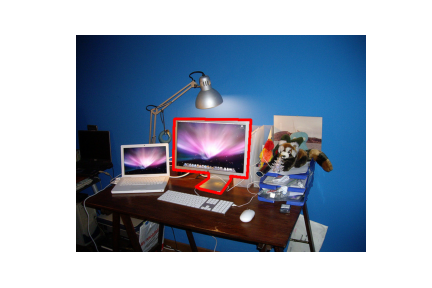}
    \put(40,0){\colorbox{white!80}{\coco{tv}}}
    \end{overpic}
    \hspace{.1cm}
    \begin{overpic}[height=2.5cm, trim={3.3cm, 1.5cm, 3.0cm, 1.4cm}, clip]
    {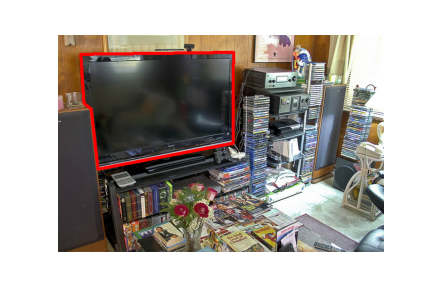}
    \put(40,0){\colorbox{white!80}{\coco{tv}}}
    \end{overpic}
    } \\ 
    \caption{Examples of instances of classes in ADE20k (in \ade{blue}) and COCO (in \coco{red}).
    The top rows shows examples for labels for which we find a relation. The second row shows how \ade{stove} and \coco{oven} categorize the visual world differently. The bottom shows different types of water which are difficult to distinguish, and different types of screens which are labeled inconsistently.}
    \label{fig:qualitative_relations}
\end{figure}
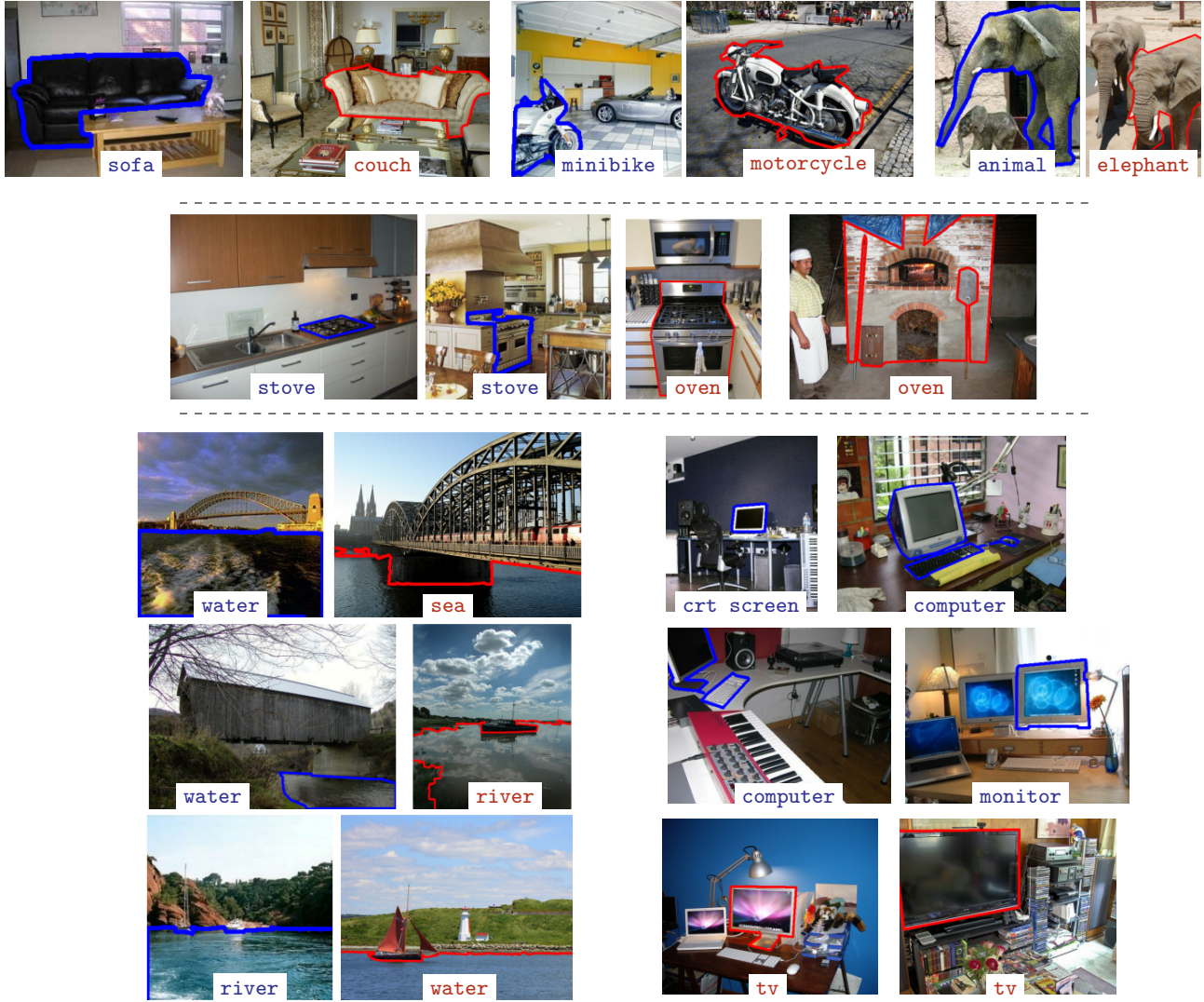

To gain insights into why and how labels relate, we visually inspect instances of labels with high-scoring relations, but whose labels do not exactly match.
We do this for relations between ADE20k and COCO which we visualize in Fig.~\ref{fig:qualitative_relations}.

\vspace{1mm}
\para{Identity.}
One of the highest scoring identity relations with non-matching labels is the ADE20k \ade{sofa} and COCO \coco{couch}. These are synonyms and indeed represent the same visual concept (see Fig.~\ref{fig:qualitative_relations} \emph{top left}). 
More interestingly, we also identify a relation between ADE20k \ade{minibike} and COCO \coco{motorcycle}. 
Semantically these are different concepts: usually a minibike denotes a tiny motorbike which is not higher than one's knees. But here both represent a full-sized motorcycle (Fig.~\ref{fig:qualitative_relations} \emph{top center}).  
Finally, another interesting, high-scoring identity relation we found is between \ade{stove} and \coco{oven} (Fig.~\ref{fig:qualitative_relations} \emph{second row}).
In ADE20k the \ade{stove} refers to the cooking panel on which you can put pots and pans, while including the oven underneath if it exists. In COCO, the \coco{oven} refers to the closed heating compartment, including the stove if it exists. 
So even while \ade{stove} and \coco{oven} are synonyms and mostly represent the same visual concept, one could argue that the true relation is not identity but overlap, because there are instances which are \ade{stove} but not \coco{oven} (2nd row, left) and vice-versa (2nd row, right).

\vspace{1mm}
\para{Parent/child.}
One example of parent/child is between ADE20k \ade{animal} and COCO \coco{elephant} (Fig.~\ref{fig:qualitative_relations} \emph{top-right}). Others include
\ade{hill} and \coco{mountain-merged},
\ade{wall} and \coco{wall-tile}.
We also correctly identify that the ADE20k \ade{tent} is a child of the COCO \coco{tent}, since the latter also includes the ADE20k \ade{awning}. 
Language alone would never be able to identify that \ade{tent} and \coco{tent} have a \emph{child} relation.

\vspace{1mm}
\para{Overlap.}
Here we look at several overlap relations found by our embeddings and logic method.
It correctly identifies the overlap between \ade{floor} and \coco{rug-merged}. This overlap relation exists because both labels use a different reference frame of the world: \ade{floor} emphasizes that the concept is \emph{stuff} and not an \emph{object}, while \coco{rug-merged} emphasizes the function and type of material (\emph{e.g.} fabric to walk on), see Fig.~\ref{fig:relation_examples}. 
We also predict an overlap relation between \ade{water} and \coco{water-other}, where the ground-truth relation is \emph{child}. When visually inspecting examples, we found many examples where it was unclear what type of water the image depicts (Fig.~\ref{fig:qualitative_relations} \emph{bottom left}). Arguably, \ade{water}, \ade{river}, \coco{sea}, \coco{river}, and \coco{water-other} all overlap, mostly caused by the visual ambiguity in images with these labels.

\vspace{1mm}
\para{Inconsistencies.}
Finally, we found strong relations not only between \ade{televi\-sion receiver} and \coco{tv}, but also between \ade{crt screen}, \ade{monitor}, \ade{computer} and \coco{tv}. Looking at instances, these labels often point to the same visual concepts (Fig.~\ref{fig:qualitative_relations} \emph{bottom right}). So strictly speaking, these labels visually overlap. However, this overlap is caused by labeling errors and inconsistencies in both datasets. In COCO, all displays (including computer monitors) are labeled as \coco{tv}.
Instead, in ADE20k computer monitors are alternatively labeled as \ade{crt screen}, \ade{monitor}, and \ade{computer}. 
These concepts overlap even within ADE20k %
which makes the common assumption of mutually exclusive labels \emph{within a dataset} invalid.

\begin{figure}[t]
    \centering
    \includegraphics[width=\textwidth]{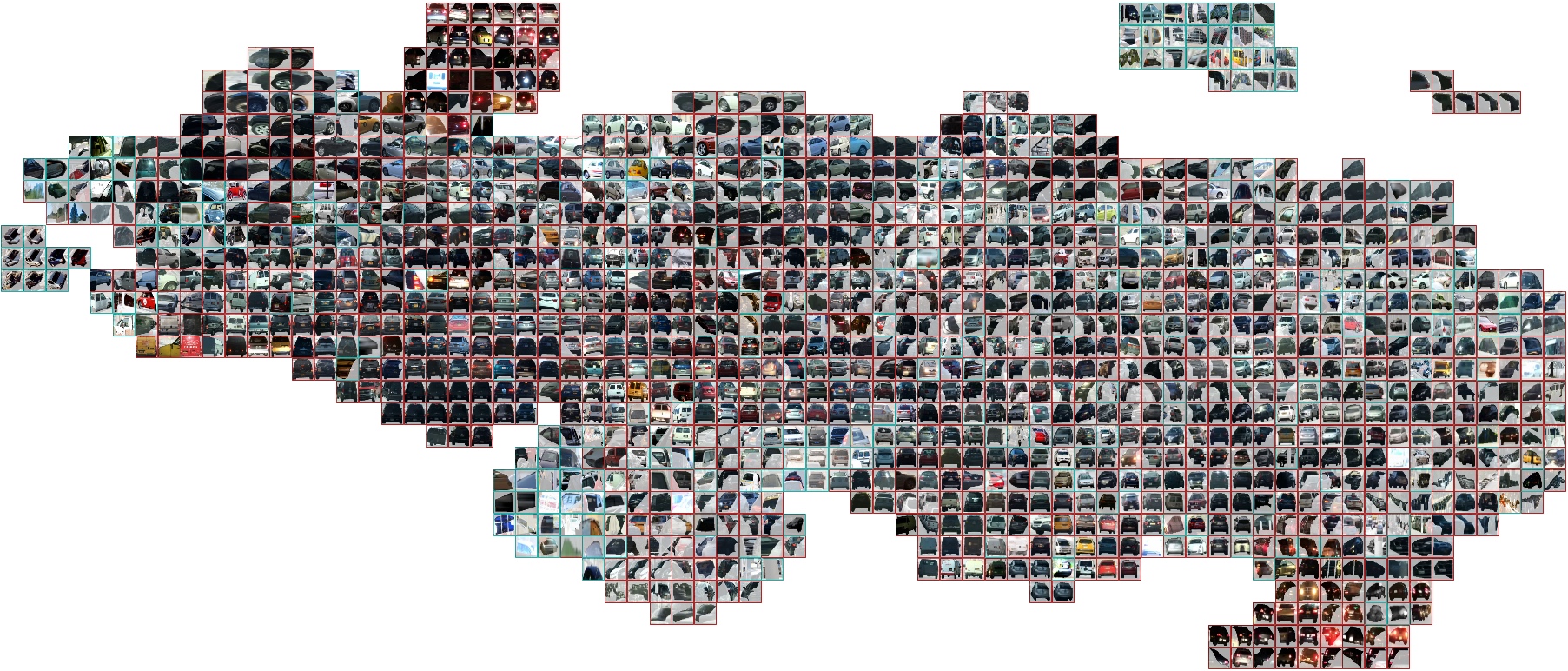}
    \vspace{-2mm}
    \caption{
        Embedding of \class{car} instances (COCO with blue box, BDD with red).
        \vspace{-2mm}
    }
    \label{fig:cocobdd_umap_instances}
\end{figure}
\begin{figure}[pt]
    \centering
    \includegraphics[width=\textwidth]{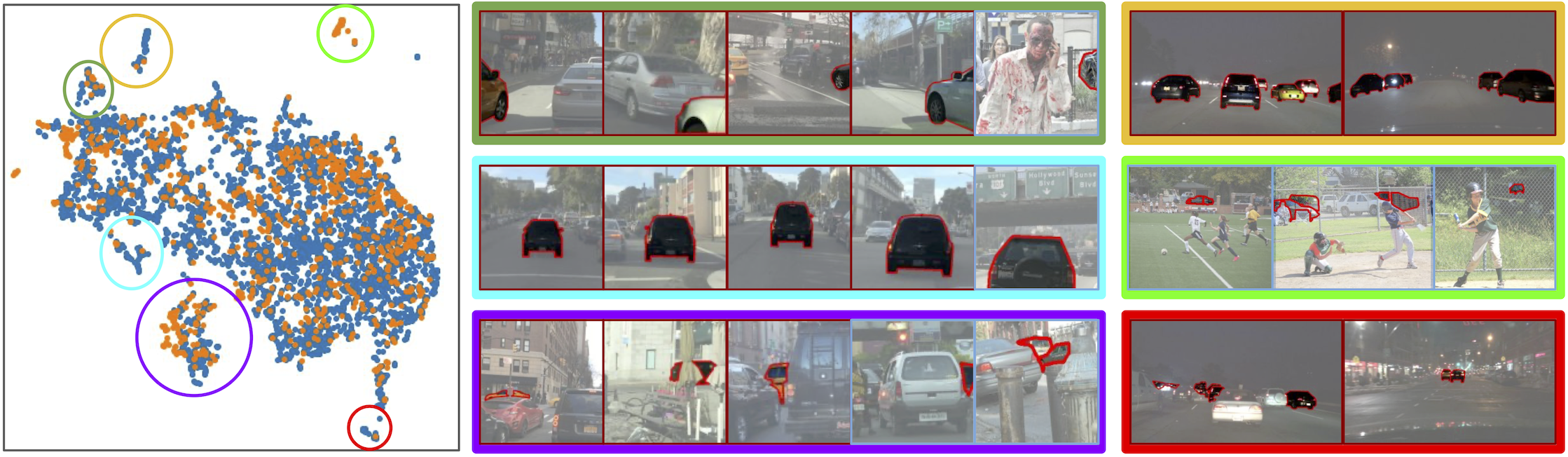}
    \vspace{-2mm}
    \caption{
    Embedding with clusters of \Coco{car} (orange dots) \& \Bdd{car} (blue dots) instances (\emph{left}) and example images sampled from each cluster (\emph{right}).
    \vspace{-2mm}
    }
    \label{fig:cocobdd_clusters}
\end{figure}

\subsection{Identify Missing Aspects}\label{sec:missing_aspects}
We want to identify which appearance aspects of a class are common between two datasets, and which are covered by only one of them.
Discovering this would enable combining examples from different datasets to cover the full range of visual appearances of a class. This could help train better recognition models.

For this experiment we focus on the \Coco{car} and \Bdd{car} classes. We use the model trained on the COCO dataset and extract features for both COCO \coco{car} and BDD \bdd{car} instances, which we aggregate per instance. 
We use these features to create a 2D visualisation using UMAP~\cite{mcinnes2018umapsoftware} in~\autoref{fig:cocobdd_umap_instances} and extract 6 different clusters for further analysis in~\autoref{fig:cocobdd_clusters}.

Each of the shown clusters has some particular visual coherence, for example:
\begin{itemize}
    \item Three clusters of cluttered streets differing in the shapes: the back of the car in the center, partial cars at the image border, and partial occluded cars.
    \item Two clusters with imagery captured at night, but with different instance shapes. Those clusters are mostly filled with images from the BDD dataset.
    \item A cluster with \emph{parked cars next to sports}, filled with only COCO images. 
\end{itemize}

\noindent From this visual analysis we observe that there is a significant overlap in the kind of \class{car} segments: both datasets contain instances with a rear or side view, partially occluded instances, and instances at the edge of the image.
However, we also find interesting differences in the imagery contained in the datasets:
BDD is a driving dataset and hence the diversity in viewpoints of scenes is limited to the viewpoint from the dashboard.
COCO, on the other hand, is a very diverse consumer dataset, where street imagery is present with much more viewpoints.
That explains why we see cars near sport fields in COCO, but not in BDD.

\begin{figure}[t]
    \centering
    \begin{subfigure}[b]{0.36\textwidth}
        \includegraphics[width=\textwidth]{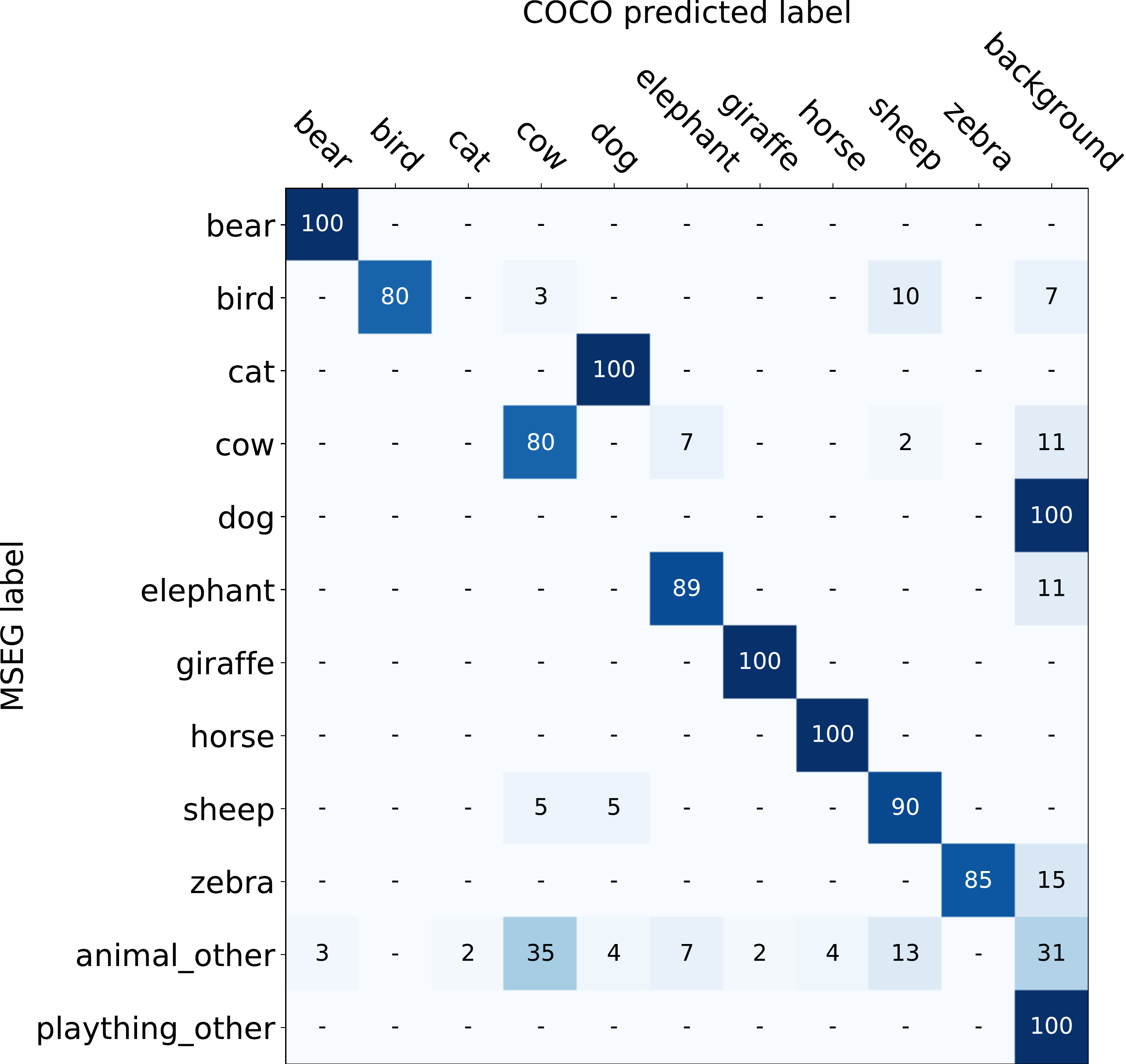}
    \end{subfigure}
    \begin{subfigure}[b]{0.63\textwidth}
        \includegraphics[width=\textwidth]{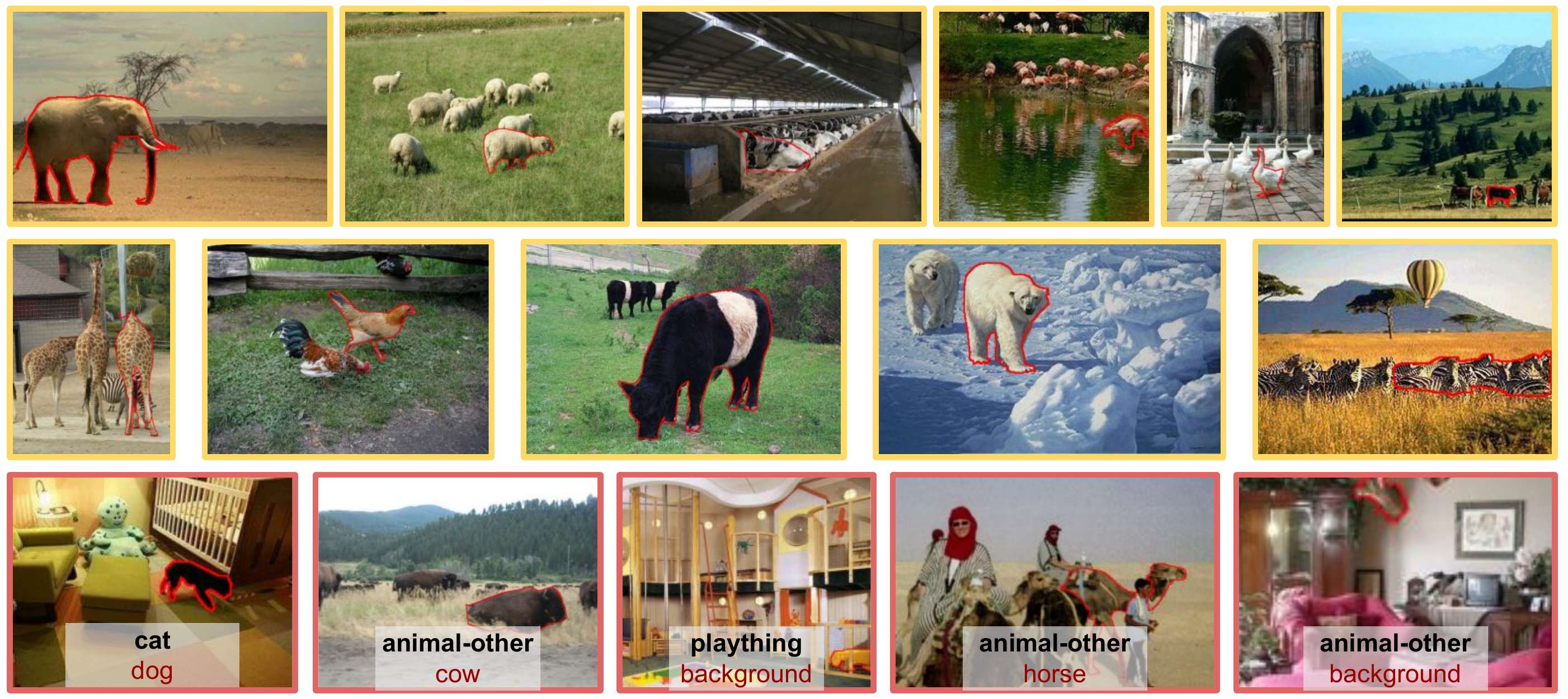}
    \end{subfigure}    
    \caption{
    Confusion matrix (\emph{left}) and example (\emph{right}) evaluating re-annotation of \Ade{animal} using the predictions of related child COCO classes.
    }
    \label{fig:ade_animals_sublabels}
\end{figure}

\subsection{Increase Label Specificity}\label{sec:label_specificity}
In this experiment we illustrate how the discovered label relations could be used to annotate images with a finer level of annotation.
Here we relabel the \Ade{animal} instances into the related COCO classes: \{\coco{cow}, \coco{dog}, \ldots, \coco{zebra}\} using the model trained on COCO, using the established label relation indicating that \Ade{animal} is a parent class of these COCO classes.

For this experiment we use the model trained on COCO and use this model to predict fine-grained annotations on the instances belonging to the \Ade{animal} class.
In order to quantitatively evaluate these new annotations we make use of the MSeg annotations.
These provide ground-truths for the ADE segments, which we use to evaluate the top-1 accuracy per class.

\autoref{fig:ade_animals_sublabels} (\emph{left}) shows the confusion matrix between MSeg ground-truths and COCO predictions on ADE instances. 
\autoref{fig:ade_animals_sublabels} (\emph{right}) shows examples of correctly and incorrectly classified segments.
From the results we observe that for most labels the finer annotations are accurate and the errors are easily explainable.

\subsection{Predict Transfer Learning Gains}\label{sec:transfer_learning}
In this section we investigate whether label relationships between two data\-sets are predictive of the gains of transfer learning.
For this we correlate the performance of transfer learning to the strength of the link between labels. 

We use our model trained on \coco{COCO} as the source model and we use \ade{ADE20k} as the target dataset.
For the label relations we use the links as discovered by the \emph{WordNet with Visual Embeddings} method.
As label link strength $s_b$ for an \ade{ADE20k} label $b$ from the \coco{COCO} dataset, we aggregate the scores over all labels $A$ in \coco{COCO} for which we have established a relation by taking the mean:
\begin{equation}
    s_b = \frac{1}{|A|} \sum_{a \in A}\ S_{a \rightarrow b}
\end{equation}

Since transfer learning is most useful when the target training set is small, we fine-tune the \coco{COCO} source model on 1000 images of the \ade{ADE20k} training set, and then evaluate per-class Intersection-Over-Union (IoU) on the (full) validation set of \ade{ADE20k}.
Following~\cite{mensink21pami}, we measure the \emph{gains} brought by transfer learning from \coco{COCO} to \ade{ADE20k} as the difference of the performance of two models:
\begin{equation}
    \textit{gains} = m_{\coco{ILSVRC12 \rightarrow COCO \rightarrow ADE20k}} - m_{\ilsvrc{ILSVRC12 \rightarrow ADE20k}}
\end{equation}

The first model performs transfer learning from \coco{COCO} to \ade{ADE20k} (after initializing the \coco{COCO} model from ILSVRC'12 as is common practice).
The second model is a baseline that trains only on \ade{ADE20k} (initialized from ILSVRC'12).
This differences measures how much transferring knowledge from \coco{COCO} helps improve performance on \ade{ADE20k}.

\begin{figure}[tb]
    \centering
    \includegraphics[width=.65\columnwidth]{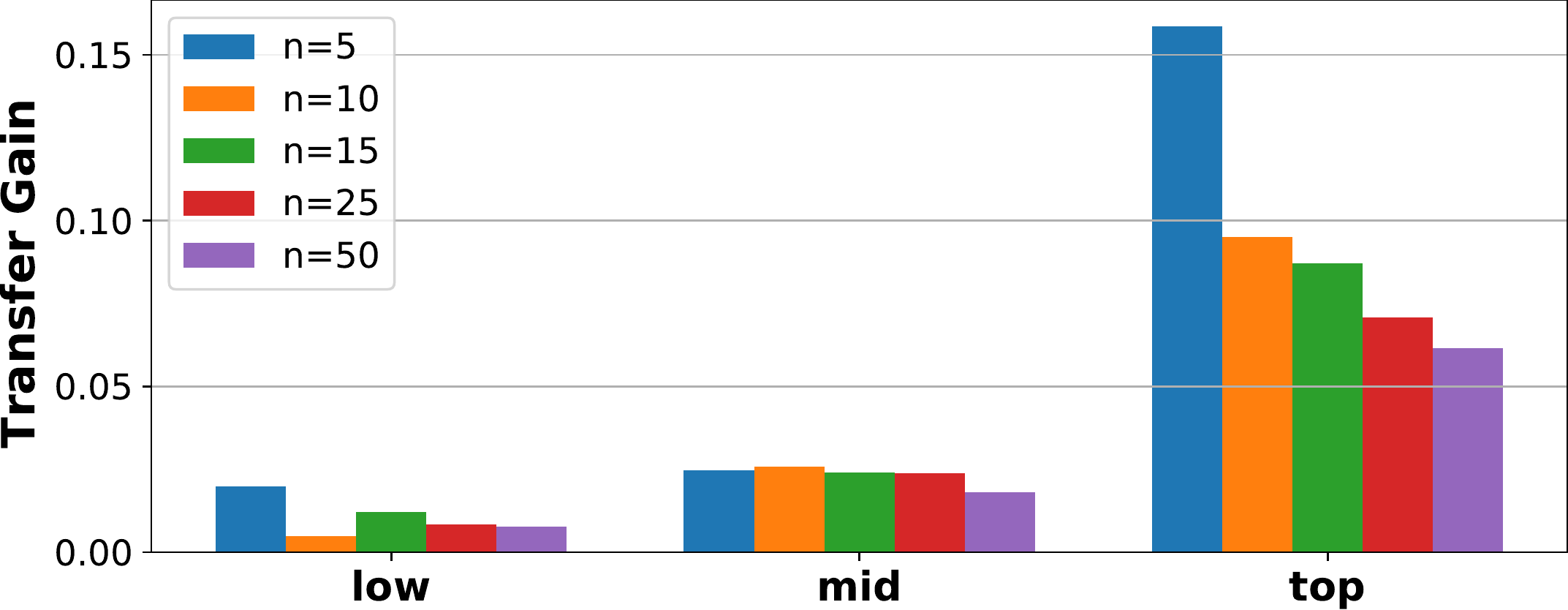}
    \caption{Mean transfer gain over labels grouped by label strength (low, mid, high). Label strength is correlated with transfer gains.}
    \label{fig:kt}
\end{figure}

In \autoref{fig:kt} we show
the performance gains averaged over the $n$ labels with the weakest label link (low), the $n$ strongest (top), and all other labels (mid). 
We observe that
(i) the mean gain over the labels with the strongest link is higher than over the labels with the weakest link;
(ii) within the top group the gain decreases as $n$ increases, and yet it remains much higher than for middle group even for $n=50$.
This indicates that labels with a stronger label link benefit more from transfer learning than labels with a weaker relation, and that is exactly what we could have expected.

Based upon these results we conclude that transfer learning indeed does bring larger gains for target labels which have a stronger link to the source dataset.

\section{Conclusion}\label{sec:conclusion}
In this paper we investigated the relations of labels across datasets.
We introduced several methods to automatically discover relations and their types.
Our experiments showed that our vision-based models outperformed our language-based models by a significant margin, demonstrating that relying on the semantics of the label names alone is insufficient for establishing such relationships. 

We demonstrated the usefulness of establishing \emph{visual-semantic} relationships on four applications. Among our findings, we discovered that the definition of labels across datasets can vary in subtle ways. Understanding these subtle relations is important when using multiple datasets, such as when training on a combination of datasets, when fine-tuning on a target dataset, or when merging two datasets.
We hope that our work inspires more researchers to study how different datasets relate to each other and how to exploit these relations to address computer vision problems.

\appendix
\section{Relate ILSVRC classification and COCO segmentation}\label{sec:ilsvrc_coco}

We now apply our visual `Pixel Probabilities' method (Sec.~\ref{sec:visual_relations}) to establish relations between labels in \coco{COCO} segmentation and \ilsvrc{ILSVRC12} classification.
We transform our \coco{COCO} segmentation model into a classification model by taking the maximum prediction score per class over all pixels in an image. We then apply it to \ilsvrc{ILSVRC12} image classification.
Vice-versa, we apply a \ilsvrc{ILSVRC12} classification model to the instances in \coco{COCO} by cropping images to each instance bounding box. As in Sec.~\ref{sec:visual_relations}, we only aggregate scores over `easy' instances to establish relations.

\begin{figure}[t!]
    \vspace{-0.3cm}
    \makebox[\linewidth][c]{%
    \begin{overpic}[height=2.5cm, trim={0.0cm, 0.0cm, 0.0cm, 0.0cm}, clip]
    {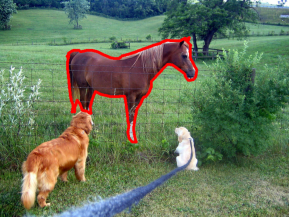}
    \put(30,0){\colorbox{white!80}{\coco{horse}}}
    \end{overpic}
    \begin{overpic}[height=2.5cm, trim={0.0cm, 0.0cm, 0.0cm, 0.0cm}, clip]
    {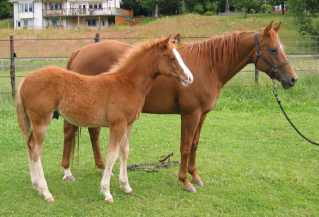}
    \put(30,0){\colorbox{white!80}{\ilsvrc{sorrel}}}
    \end{overpic}
    \begin{overpic}[height=2.5cm, trim={0.0cm, 0.0cm, 0.0cm, 0.0cm}, clip]
    {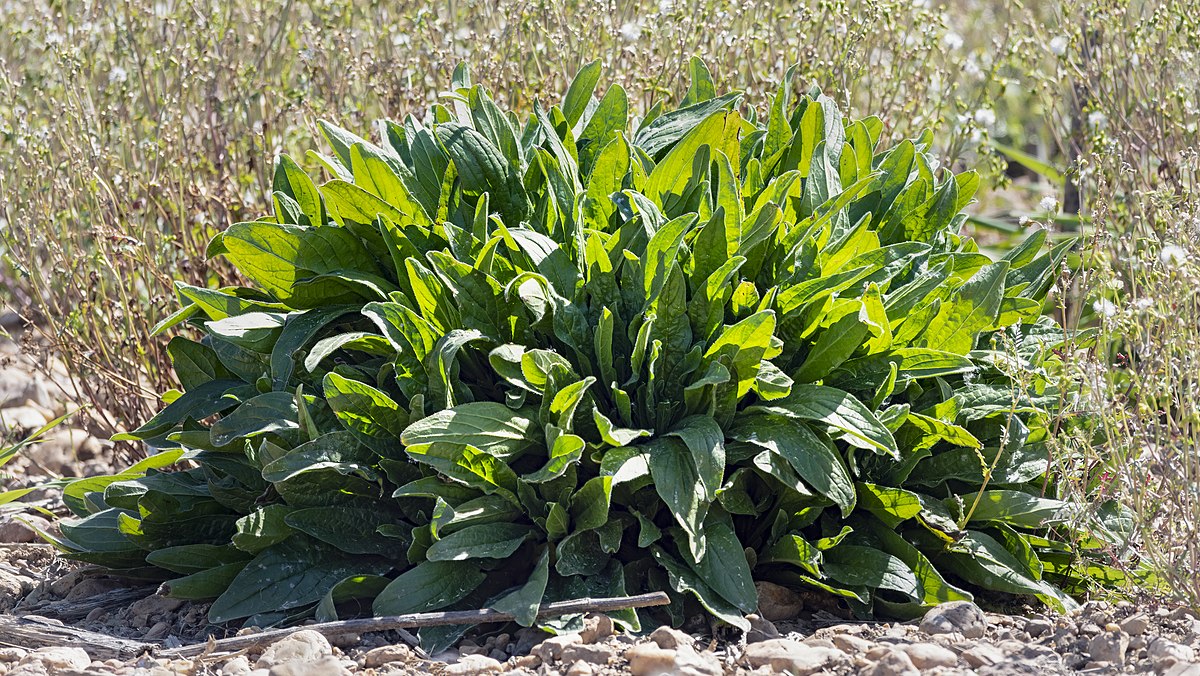}
    \put(10,0){\colorbox{white!80}{\texttt{sorrel (web search)}}}
    \end{overpic}
    } 
    \vspace{-.1cm}
    \begin{tikzpicture}
    \draw[dashed] (-10,0) -- (3,0);
    \end{tikzpicture}
    \vspace{-.1cm}
    
    \makebox[\linewidth][c]{%
    \begin{overpic}[height=2.5cm, trim={0.0cm, 0.0cm, 0.0cm, 0.0cm}, clip]
    {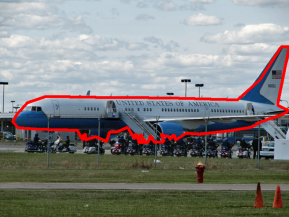}
    \put(25,0){\colorbox{white!80}{\coco{airplane}}}
    \end{overpic}
    \begin{overpic}[height=2.5cm, trim={0.0cm, 0.0cm, 0.0cm, 0.0cm}, clip]
    {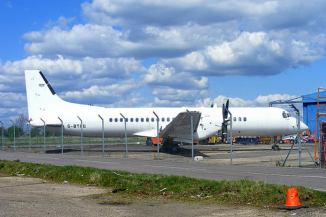}
    \put(40,0){\colorbox{white!80}{\ilsvrc{wing}}}
    \end{overpic}
    \begin{overpic}[height=2.5cm, trim={0.0cm, 0.0cm, 0.0cm, 0.0cm}, clip]
    {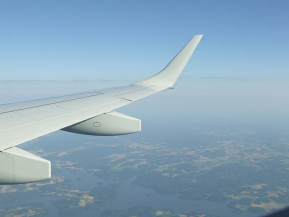}
    \put(40,0){\colorbox{white!80}{\ilsvrc{wing}}}
    \end{overpic}
    } 
    \vspace{-.1cm}
    \begin{tikzpicture}
    \draw[dashed] (-10,0) -- (3,0);
    \end{tikzpicture}
    \vspace{-.1cm}

    \makebox[\linewidth][c]{%
    \begin{overpic}[height=2.5cm, trim={4.0cm, 0.0cm, 0.0cm, 0.0cm}, clip]
    {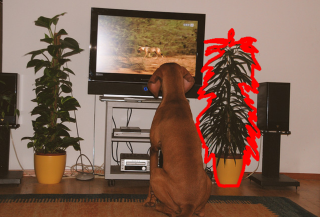}
    \put(2,0){\colorbox{white!80}{\coco{potted plant}}}
    \end{overpic}
    \begin{overpic}[height=2.5cm, trim={0.0cm, 0.0cm, 0.0cm, 0.0cm}, clip]
    {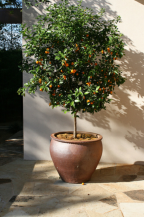}
    \put(20,0){\colorbox{white!80}{\ilsvrc{pot}}}
    \end{overpic}
    \begin{overpic}[height=2.5cm, trim={0.0cm, 0.0cm, 0.0cm, 0.0cm}, clip]
    {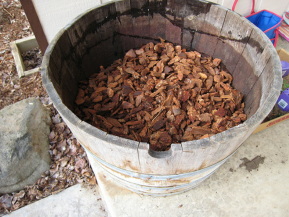}
    \put(40,0){\colorbox{white!80}{\ilsvrc{pot}}}
    \end{overpic}
    \begin{overpic}[height=2.5cm, trim={0.0cm, 0.0cm, 0.0cm, 0.0cm}, clip]
    {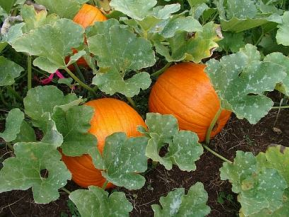}
    \put(40,0){\colorbox{white!80}{\ilsvrc{pot}}}
    \end{overpic}
    \hspace{3mm}
    \begin{overpic}[height=2.5cm, trim={2.0cm, 0.0cm, 2.0cm, 0.0cm}, clip]
    {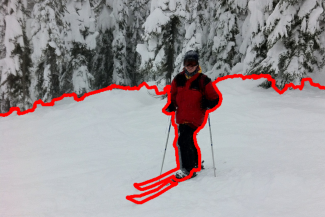}
    \put(40,0){\colorbox{white!80}{\coco{snow}}}
    \end{overpic}
    \begin{overpic}[height=2.5cm, trim={0.0cm, 0.0cm, 0.0cm, 0.0cm}, clip]
    {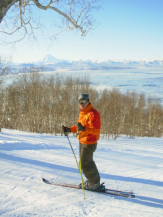}
    \put(20,0){\colorbox{white!80}{\ilsvrc{ski}}}
    \end{overpic}
    } \\ 
    \caption{Examples of instances of classes in COCO (in \coco{red}) and ILSVRC12 (in \ilsvrc{purple}) for which we find a relation. See main text for details.}
    \label{fig:ilsvrc_coco}
\end{figure}
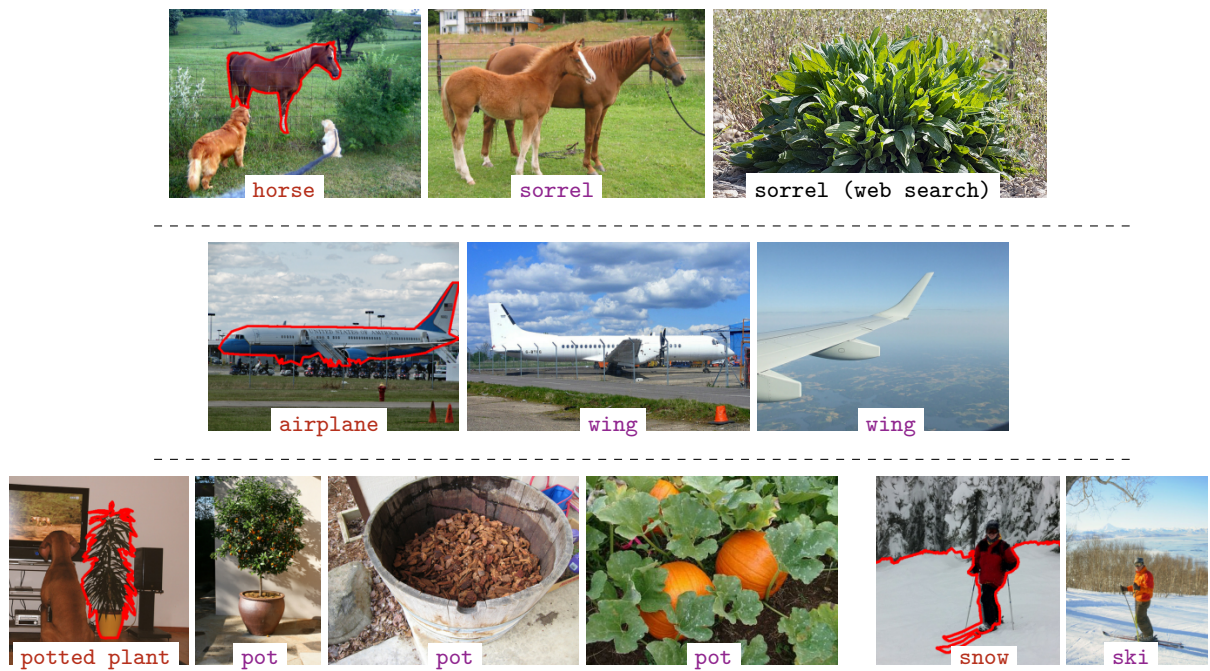

Since for this experiment we do not have the ground truth relations, we manually inspect the top 100 relations predicted by our method. We found that 80\% of them are correct. Not all relations can be found based on language alone. For example, we found that COCO's \coco{horse} is related to ILSVRC12's \ilsvrc{sorrel} (a type of horse, while `sorrel'  commonly refers to a plant, see Fig~\ref{fig:ilsvrc_coco} \emph{top row}).
In 9\% of the cases, the label names suggest they are in a part-of relation.
However, inspecting the visual examples reveals that in some cases this is not true.
For example, we predict \coco{toilet} - \ilsvrc{toilet\_seat} to be in an identity relation. In fact, most \ilsvrc{toilet\_seats} are full toilets (only 5\% \ilsvrc{toilet\_seat} with no toilet, and even ~4\% \ilsvrc{toilet\_seat} without seat).
Another example is \coco{potted plant} - \ilsvrc{pot}. Again, most \ilsvrc{pots} contain a plant, with 8\% depicting only a plant, and 5\% only a pot (Fig.~\ref{fig:ilsvrc_coco} \emph{bottom right}).
Finally, we found \coco{airplane} - \ilsvrc{wing}, where the latter is indeed an airplane wing, not an animal wing. Here 25\% of the \ilsvrc{wing} images depict a full airplane (Fig.~\ref{fig:ilsvrc_coco} \emph{middle row}).

The remaining 11\% of predicted relations are wrong, often due to contextual errors. For example, we predict \coco{snow} - \ilsvrc{ski} (Fig.~\ref{fig:ilsvrc_coco} \emph{bottom right}). Such mistakes may be avoidable by using COCO masks instead of boxes or by using language priors.

Finally, we investigated all \coco{dog} - \ilsvrc{ilsvrc} relations. Our method predicts 159 such relations, all as type \emph{parent-of}. Remarkably, all 118 finegrained ILSVRC dog labels are included in the highest scored 129 relations we discovered.

We conclude that our method can work across different types of datasets.

\bibliographystyle{splncs04}
\bibliography{shortstrings,loco,loco_extra}

\begin{thebibliography}{10}
\providecommand{\url}[1]{\texttt{#1}}
\providecommand{\urlprefix}{URL }
\providecommand{\doi}[1]{https://doi.org/#1}

\bibitem{robustvisionchallenge}
Robust vision challenge. \url{http://www.robustvision.net/}

\bibitem{bevandic22wacv}
Bevandi\'c, P., Or\v{s}i\'c, M., Grubi\v{s}i\'c, I., \v{S}ari\'c, J.,
  \v{S}egvi\'c, S.: Multi-domain semantic segmentation with overlapping labels.
  In: Proc. WACV (2022)

\bibitem{bucher19neurips}
Bucher, M., Vu, T., Cord, M., P{\'{e}}rez, P.: Zero-shot semantic segmentation.
  In: NeurIPS (2019)

\bibitem{cocostuff-dataset}
Caesar, H., Uijlings, J., Ferrari, V.: {COCO-Stuff} dataset.
  \url{http://calvin.inf.ed.ac.uk/datasets/coco-stuff} (2018)

\bibitem{caesar18cvpr}
Caesar, H., Uijlings, J., Ferrari, V.: {COCO-Stuff}: Thing and stuff classes in
  context. In: CVPR (2018)

\bibitem{deng09cvpr}
Deng, J., Dong, W., Socher, R., Li, L.J., Li, K., Fei-fei, L.: {ImageNet}: {A}
  large-scale hierarchical image database. In: CVPR (2009)

\bibitem{everingham15ijcv}
Everingham, M., Eslami, S., van Gool, L., Williams, C., Winn, J., Zisserman,
  A.: The {PASCAL} visual object classes challenge: A retrospective. IJCV
  (2015)

\bibitem{geiger13ijrr}
Geiger, A., Lenz, P., Stiller, C., Urtasun, R.: Vision meets robotics: The
  {KITTI} dataset. International Journal of Robotics Research  (2013)

\bibitem{ghiasi21arxiv}
Ghiasi, G., Gu, X., Cui, Y., Lin, T.: Open-vocabulary image segmentation. Tech.
  rep., ArXiV (2021)

\bibitem{word2vectfhub}
Google: Wiki words 500 with normalization - a 500 dimensional wor2vec skip-gram
  model trained on english wikipedia.
  \url{https://tfhub.dev/google/Wiki-words-500-with-normalization/2}

\bibitem{jia21icml_align}
Jia, C., Yang, Y., Xia, Y., Chen, Y., Parekh, Z., Pham, H., Le, Q.V., Sung, Y.,
  Li, Z., Duerig, T.: Scaling up visual and vision-language representation
  learning with noisy text supervision. In: ICML (2021)

\bibitem{kirillov18coco}
Kirillov, A.: Panoptic challenge intro. COCO+Mapillary Joint Recognition
  Challenge Workshop,
  \url{http://presentations.cocodataset.org/ECCV18/COCO18-Panoptic-Overview.pdf}

\bibitem{kirillov19cvpr}
Kirillov, A., He, K., Girshick, R., Rother, C., Doll\'ar, P.: Panoptic
  segmentation. In: CVPR (2019)

\bibitem{kokkinos17cvpr}
Kokkinos, I.: Ubernet: Training a ‘universal’ cnn for low-, mid-, and high-
  level vision using diverse datasets and limited memory. In: CVPR (2017)

\bibitem{kuznetsova20ijcv}
Kuznetsova, A., Rom, H., Alldrin, N., Uijlings, J., Krasin, I., Pont-Tuset, J.,
  Kamali, S., Popov, S., Malloci, M., Duerig, T., Ferrari, V.: {The Open Images
  Dataset V4: Unified image classification, object detection, and visual
  relationship detection at scale}. IJCV  (2020)

\bibitem{lambert20cvpr}
Lambert, J., Liu, Z., Sener, O., Hays, J., Koltun, V.: {MSeg}: A composite
  dataset for multi-domain semantic segmentation. In: CVPR (2020)

\bibitem{lin14eccv}
Lin, T.Y., Maire, M., Belongie, S., Bourdev, L., Girshick, R., Hays, J.,
  Perona, P., Ramanan, D., Zitnick, C.L., Doll\'{a}r, P.: Microsoft {COCO}:
  Common objects in context. In: ECCV (2014)

\bibitem{mcinnes2018umapsoftware}
McInnes, L., Healy, J., Saul, N., Grossberger, L.: {UMAP}: Uniform manifold
  approximation and projection. The Journal of Open Source Software  (2018)

\bibitem{mensink21pami}
Mensink, T., Uijlings, J., Kuznetsova, A., Gygli, M., Ferrari, V.: Factors of
  influence for transfer learning across diverse appearance domains and task
  types. IEEE Trans. on PAMI  (2021)

\bibitem{mikolov13iclr}
Mikolov, T., Chen, K., Corrado, G., Dean, J.: Efficient estimation of word
  representations in vector space. In: ICLR workshop (2013)

\bibitem{miller95acm}
Miller, G.: { WordNet: a lexical database for English }. Communications of the
  ACM  \textbf{38}(11),  39--41 (1995)

\bibitem{ponce06datasetissues}
Ponce, J., Berg, T.L., Everingham, M., Forsyth, D.A., Hebert, M., Lazebnik, S.,
  Marszalek, M., Schmid, C., Russell, B.C., Torralba, A., Williams, C.K.I.,
  Zhang, J., Zisserman, A.: Dataset issues in object recognition. In: Toward
  Category-Level Object Recognition (2006)

\bibitem{radford21icml_clip}
Radford, A., Kim, J.W., Hallacy, C., Ramesh, A., Goh, G., Agarwal, S., Sastry,
  G., Askell, A., Mishkin, P., Clark, J., Krueger, G., Sutskever, I.: Learning
  transferable visual models from natural language supervision. In: ICML (2021)

\bibitem{rebuffi17nips}
Rebuffi, S.A., Bilen, H., Vedaldi, A.: Learning multiple visual domains with
  residual adapters. In: NeurIPS (2017)

\bibitem{silberman12eccv}
Silberman, N., Hoiem, D., Kohli, P., Fergus, R.: Indoor segmentation and
  support inference from rgbd images. In: ECCV (2012)

\bibitem{Torralba11}
Torralba, A., Efros, A.: An unbiased look on dataset bias. In: CVPR (2011)

\bibitem{triantafillou20iclr}
Triantafillou, E., Zhu, T., Dumoulin, V., Lamblin, P., Evci, U., Xu, K.,
  Goroshin, R., Gelada, C., Swersky, K., Manzagol, P., Larochelle, H.:
  Meta-dataset: {A} dataset of datasets for learning to learn from few
  examples. In: ICLR (2020)

\bibitem{wang20pami}
{Wang}, J., {Sun}, K., {Cheng}, T., {Jiang}, B., {Deng}, C., {Zhao}, Y., {Liu},
  D., {Mu}, Y., {Tan}, M., {Wang}, X., {Liu}, W., {Xiao}, B.: Deep
  high-resolution representation learning for visual recognition. IEEE Trans.
  on PAMI  (2020)

\bibitem{xiao10cvpr}
Xiao, J., Hays, J., Ehinger, K., Oliva, A., Torralba, A.: {SUN} database:
  Large-scale scene recognition from {A}bbey to {Z}oo. In: CVPR (2010)

\bibitem{xiao13iccv}
Xiao, J., Owens, A., Torralba, A.: {SUN3D}: A database of big spaces
  reconstructed using {SfM} and object labels. In: ICCV (2013)

\bibitem{fisher20cvpr}
Yu, F., Chen, H., Wang, X., Xian, W., Chen, Y., Liu, F., Madhavan, V., Darrell,
  T.: Bdd100k: A diverse driving dataset for heterogeneous multitask learning.
  In: CVPR (2020)

\bibitem{yu20cvpr}
Yu, F., Chen, H., Wang, X., Xian, W., Chen, Y., Liu, F., Madhavan, V., Darrell,
  T.: Bdd100k: A diverse driving dataset for heterogeneous multitask learning.
  In: CVPR (2020)

\bibitem{zendel17cvpr}
Zendel, O., Honauer, K., Murschitz, M., Humenberger, M., Fernandez~Dominguez,
  G.: Analyzing computer vision data - the good, the bad and the ugly. In: CVPR
  (2017)

\bibitem{zhou17cvpr}
Zhou, B., Zhao, H., Puig, X., Fidler, S., Barriuso, A., Torralba, A.: Scene
  parsing through {ADE20K} dataset. In: CVPR (2017)

\bibitem{zhou14nips}
Zhou, B., Lapedriza, A., Xiao, J., Torralba, A., Oliva, A.: Learning deep
  features for scene recognition using places database. In: NeurIPS (2014)

\bibitem{zhou22cvpr}
Zhou, X., Koltun, V., Kr{\"{a}}henb{\"{u}}hl, P.: Simple multi-dataset
  detection. In: CVPR (2022)

\end{thebibliography}

\end{document}